\documentclass[manuscript]{acmart}
\usepackage{tikz}
\usepackage{multicol}
\usepackage{multirow}
\usepackage{tabularx}
\usepackage{todonotes}
\usepackage{xcolor}
\usepackage{menukeys}
\usepackage[ruled,vlined]{algorithm2e}
\usepackage{multirow}
\usepackage[flushleft]{threeparttable}
\usepackage{circledsteps}
\usepackage{threeparttable}
\usepackage{hyperref}
\usepackage[capitalise,nameinlink]{cleveref}
\usepackage{booktabs}
\usepackage{multirow}
\usepackage{tabularx}
\usepackage{graphicx}
\usepackage{caption}
\usepackage{xspace}
\usepackage{subcaption}
\usepackage[normalem]{ulem}
\usepackage{tcolorbox}
\usepackage{xurl}
\usepackage{algorithmic}
\AtBeginDocument{%
  }

\setcopyright{acmlicensed}
\copyrightyear{2026}
\acmYear{2026}
\acmDOI{XXXXXXX.XXXXXXX}

\acmJournal{HEALTH}
\acmVolume{0}
\acmNumber{0}
\acmArticle{0}
\acmMonth{0}

\begin{document}

\title{End-to-end Automated Deep Neural Network Optimization for PPG-based Blood Pressure Estimation on Wearables}
\author{Francesco Carlucci}
\email{francesco.carlucci@polito.it}
\orcid{}
\affiliation{%
  \institution{Interuniversity Department of Regional and Urban Studies and Planning at Politecnico di Torino}
  \city{Torino}
  \country{Italy}
}
\author{Giovanni Pollo}
\email{giovanni.pollo@polito.it}
\orcid{}
\affiliation{%
  \institution{Department of Control and Computer Engineering at Politecnico di Torino}
  \city{Torino}
  \country{Italy}
}
\author{Xiaying Wang}
\email{}
\orcid{}
\affiliation{%
  \institution{Department of Information Technology and Electrical Engineering at ETH Zurich}
  \city{Zurich}
  \country{Switzerland}
}
\author{Massimo Poncino}
\email{massimo.poncino@polito.it}
\orcid{}
\affiliation{%
  \institution{Department of Control and Computer Engineering at Politecnico di Torino}
  \city{Torino}
  \country{Italy}
}
\author{Enrico Macii}
\email{enrico.macii@polito.it}
\orcid{}
\affiliation{%
  \institution{Interuniversity Department of Regional and Urban Studies and Planning at Politecnico di Torino}
  \city{Torino}
  \country{Italy}
}
\author{Luca Benini}
\email{lbenini@iis.ee.ethz.ch}
\orcid{}
\affiliation{%
  \institution{Department of Information Technology and Electrical Engineering at ETH Zurich}
  \city{Zurich}
  \country{Switzerland}
}
\affiliation{%
  \institution{Department of Electrical, Electronic, and Information Engineering "Guglielmo Marconi" at University of Bologna}
  \city{Bologna}
  \country{Italy}
}
\author{Sara Vinco}
\email{sara.vinco@polito.it}
\orcid{}
\affiliation{%
  \institution{Department of Control and Computer Engineering at Politecnico di Torino}
  \city{Torino}
  \country{Italy}
}
\author{Alessio Burrello}
\email{alessio.burrello@polito.it}
\orcid{}
\affiliation{%
  \institution{Department of Control and Computer Engineering at Politecnico di Torino}
  \city{Torino}
  \country{Italy}
}

\author{Daniele Jahier Pagliari}
\email{daniele.jahier@polito.it}
\orcid{}
\affiliation{%
  \institution{Department of Control and Computer Engineering at Politecnico di Torino}
  \city{Torino}
  \country{Italy}
}
\renewcommand{\shortauthors}{Carlucci, Pollo et al.}
\acmArticleType{Review}
\acmCodeLink{https://github.com/borisveytsman/acmart}
\acmDataLink{htps://zenodo.org/link}
\acmContributions{BT and GKMT designed the study; LT, VB, and AP
  conducted the experiments, BR, HC, CP and JS analyzed the results,
  JPK developed analytical predictions, all authors participated in
  writing the manuscript.}

\newcommand{\rev}[1]{\textcolor{black}{#1}}

\begin{abstract}
Photoplethysmography (PPG)-based blood pressure (BP) estimation is a challenging task, particularly on resource-constrained wearable devices. However, fully on-board processing is desirable to ensure user data confidentiality. Recent deep neural networks (DNNs) have achieved high BP estimation accuracy by reconstructing BP waveforms or directly regressing BP values, but their large memory, computation, and energy requirements hinder deployment on wearables. This work introduces a fully automated DNN design pipeline that combines hardware-aware neural architecture search (NAS), pruning, and mixed-precision search (MPS) to generate accurate yet compact BP prediction models optimized for ultra-low-power multicore systems-on-chip (SoCs). Starting from state-of-the-art baseline models on four public datasets, our optimized networks achieve up to 7.99\% lower error with a 7.5$\times$ parameter reduction, or up to 83$\times$ fewer parameters with negligible accuracy loss. All models fit within 512 kB of memory on our target SoC (GreenWaves' GAP8), requiring less than 55 kB and achieving an average inference latency of 142 ms and energy consumption of 7.25 mJ. Patient-specific fine-tuning further improves accuracy by up to 64\%, enabling fully autonomous, low-cost BP monitoring on wearables.
\end{abstract}

\maketitle
\section{Introduction}
Blood Pressure (BP) is a critical health parameter. 
It is one of the most significant risk factors for Cardiovascular Diseases (CVDs)\cite{doi:10.1161/HYPERTENSIONAHA.119.14240}, including High Blood Pressure, atrial fibrillation, myocardial infarction, and aortic rupture \cite{ctx2426934390007866, Odutayoi4482}. These conditions contribute to a high number of deaths worldwide \cite{PMID:31222515}. Consequently, periodic or continuous BP monitoring is highly desirable for early diagnosis and prevention~\cite{Geerse2019}.

BP monitoring techniques can be broadly classified into two categories: invasive and non-invasive methods \cite{Ward2007, 9268175, Meidert2017}. Among invasive methods, arterial cannulation is the most accurate. This technique, which directly measures Atrial Blood Pressure (ABP), involves inserting a thin catheter with a cannula directly into an artery. However, due to the necessity of specialized equipment and trained personnel, it is predominantly used in hospital settings for precise BP monitoring.

Among non-invasive methods, cuff-based sphygmomanometers are the most widely used and accurate. These devices consist of a pressure cuff wrapped around the upper arm, an air pump, and a manometer. By inflating the cuff to compress the artery and gradually releasing the pressure, BP is measured with high reliability. However, despite their accuracy and non-invasiveness, cuff-based methods are not suitable for continuous monitoring, as they require specialized equipment uncomfortable to wear during daily activities and a controlled environment.

A promising alternative that enables both non-invasive and continuous BP monitoring is Photoplethysmography (PPG). PPG is an optical technique that measures blood volume changes in the skin using light-emitting diodes (LEDs) and photodetectors. This method has gained significant popularity due to its integration into consumer wearable devices such as smartwatches and fitness bands. An LED sensor illuminates the skin, and a photodiode captures the reflected light, the intensity of which varies according to blood volume fluctuations caused by heart activity~\cite{ppg-signal}.

Other medically relevant parameters can be derived from PPG with high accuracy, such as heart rate (HR) \cite{benfenati2024enhanceppgimprovingppgbasedheart, ppg-hr} and respiratory rate \cite{9176231}.
However, in this paper, we focus on its usage for the estimation of Systolic Blood Pressure (SBP) and Diastolic Blood Pressure (DBP), which is an actively explored field, with accuracies not reaching medical grade.

Given the PPG signal, estimating the SBP and DBP is non-trivial. 
Different techniques have already been explored, including Machine Learning (ML) methods that rely on hand-crafted features, such as Pulse Transit Time (PTT) \cite{7118672}, as well as Deep Learning (DL) approaches that learn directly from raw or minimally processed signals \cite{s20195606, ppg-dnn-1, ppg-dnn-2}.
Among ML methods, techniques such as a Random Forest (RF)~\cite{ppg-rf} or an ensemble of Support Vector Regression (SVR)~\cite{ppg-svr} trained with bagging, have been explored. Compared to traditional ML-based methods, Deep Neural Networks (DNNs) offer the advantage of bypassing the often costly feature extraction process and have demonstrated superior generalization capabilities for unseen data in biosignal processing tasks\cite{q-ppg, burrello2022bioformers, ppg-tcn}. Different DNN architectures have been investigated for PPG-based BP estimation\cite{ppg-dnn-1,ppg-dnn-2,ppg-mlp,ppg-lstm}, with recent research focusing on 1D Convolutional Neural Networks (CNNs)\cite{ppg-tcn}.

However, existing deep learning models for BP estimation have a large number of parameters and high computational complexity \cite{arjomand2025transforhythmtransformerarchitectureconductive, 10568937, Tian2025}. When pursuing continuous monitoring on resource-constrained, low-power devices such as wearables, those models either exceed the available memory or incur excessive latency~\cite{10798404}.
As a result commercial wearable BP systems \cite{almeidaAktiia2023} adopt remote processing pipelines, where raw data is collected and processed on the cloud. This approach is non-real time and raises concerns on data privacy and security \cite{rt-ppg-id, ppg_auth}.

This paper, which extends the preliminary work presented in \cite{10798404}, aims at reducing the complexity of DNN-based methods while maintaining their high accuracy, through the use of a fully automated hardware-aware DNN compression pipeline. The starting points of our optimization are DNN architectures derived by a previous study~\cite{nature-survey}, representing the current state-of-the-art (SotA) for the task.
The following are the main contributions of this work:

\begin{itemize}
    \item We propose a fully automated DNN optimization toolchain for PPG-based BP estimation:
    first, we leverage a \textbf{gradient-based Neural Architecture Search (NAS)} to automatically select DNN layers from a predefined pool and optimize network depth.
    Second, we apply \textbf{structured pruning} to eliminate redundant portions of the convolutional and fully connected layers.
    Lastly, we apply \textbf{mixed precision quantization}, obtaining integer deployable models that utilize \texttt{int2}, \texttt{int4}, \texttt{int8} data types for both weights and activation, further reducing memory occupation.
    \item On four different open-source datasets \cite{sensors_dataset, uci_dataset, BCG_dataset, PPGB_dataset}, our integrated pipeline produces models with up to 7.99\% lower error compared to the best DL SotA with a 7.5$\times$ parameter reduction, or with up to 83$\times$ fewer parameters with negligible accuracy loss.
    \item We deploy Pareto-optimal models, \rev{i.e., models that balance model size and estimation error optimally, among those found with our optimizations} on a SotA multi-core RISC-V ultra-low-power System on Chip (SoC), GAP8 \cite{gap8}, exploiting an open-source deployment tool, DORY \cite{dory}. 
    On the BCG Dataset \cite{BCG_dataset}, i.e. the one with the most samples per patient among the four considered, all our deployed models occupy less than 55 kB, while reaching an average latency per inference of 142.14 ms (min: 52.87, max: 241.33) and a corresponding energy consumption of 7.25 mJ (min: 2.70, max: 12.31).
    \item As a further experiment, we investigate \textbf{patient-specific fine-tuning} to improve the accuracy of our DNNs. Using a consistent and leakage-free split strategy, fine-tuning yields accuracy improvements of up to 61.1\% for SBP and 64.27\% for DBP. 
    \rev{Although the study cohort ($n = 40$) is smaller than the 85-subject minimum required by the AAMI protocol \cite{White1993}, a widely used validation standard for BP measurement devices that prescribes both a minimum cohort size and acceptance thresholds to assess agreement with reference measurements, our most accurate models satisfy its core accuracy criterion (mean error $\leq$ 5 mmHg and standard deviation $\leq$ 8 mmHg). Notably, the best-performing of the three deployed models achieved a Mean Error (ME) of 1.39 mmHg with a Standard Deviation (STD) of 2.36 mmHg on DBP.}

\end{itemize}

\section{Background and Related Works}

\begin{figure}
    \centering
    \includegraphics[width=0.9\linewidth]{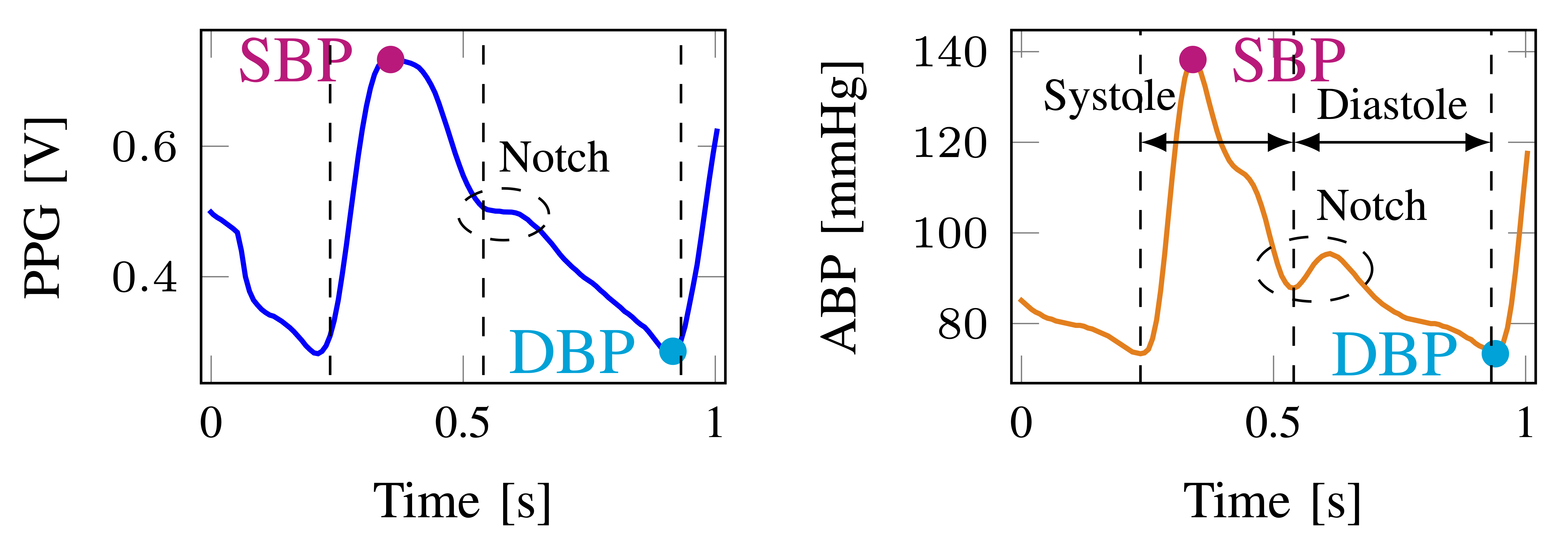}
    \caption{\rev{SBP and DBP estimation from PPG (left) and ABP (right) signals.}}
    \label{fig:BP_PPG}
\end{figure}

The study of BP monitoring solutions based on wearable devices equipped with PPG sensors has attracted significant research interest from both academia and industry in recent years.

\rev{Figure \ref{fig:BP_PPG} compares the PPG and ABP waveforms over a single cardiac cycle. In both signals, when the left ventricle contracts and pumps blood into the arteries (Systole), the waveforms rise and reach the Systolic Blood Pressure (SBP), corresponding to the maximum arterial pressure. After this push, the aortic valve closes and the heart relaxes; the pressure then slowly drops as blood keeps flowing through the arteries to the rest of the body. The lowest point before the next beat is the Diastolic Blood Pressure (DBP). A small dip called the dicrotic notch can be observed around the moment the aortic valve closes, marking the transition from Systole to Diastole. 
This shared morphology helps explain why PPG can contain informative cues for BP estimation \cite{correlation_study}. However, practical BP estimation from PPG in daily wearable use remains challenging, because the optical signal is sensitive to various types of artifacts due to noise, changes at the skin-sensor interface, etc~\cite{10.1145/3587256}.}

Among the proposed solutions, two main modeling approaches have emerged: signal-to-label models and signal-to-signal models. The first directly estimates discrete SBP and DBP values from segments of the PPG signal. In contrast, signal-to-signal models reconstruct the continuous ABP waveform from the PPG signal. The values of systolic and diastolic blood pressures are then measured from the peaks and valleys of such a reconstructed signal.

\rev{Researchers typically measure BP estimation accuracy using the Mean Absolute Error (MAE) for both SBP and DBP, calculated as $MAE_{SBP} = \mathbb{E}(|SBP_{true} - SBP_{pred}|)$ and $MAE_{DBP} = \mathbb{E}(|DBP_{true} - DBP_{pred}|)$ respectively, where $\mathbb{E}()$ represents the average over all predictions in the test set}. Recognizing the importance of standardized validation, the IEEE established guidelines in 2014 for wearable cuffless BP devices, which mandate reporting MAE as a key performance metric in validation results \cite{6882122}.

\rev{Table \ref{tab:related-works-others} and Table \ref{tab:related-works-benchmark}} summarize the major recent approaches for PPG-based BP estimation, focusing on ML and DL methods. A key difficulty when analyzing state-of-the-art methods for this task lies in the fact that many of them use custom preprocessing and dataset preparation techniques (input filtering, data splitting, etc), often unrealistic in practice (e.g., filtering out particularly noisy records without a well-grounded biological “threshold”) or introducing various degrees of information leakage. This makes a direct numerical comparison between MAE results of little meaning. Therefore, we categorize related works into two main groups:
i) \textit{Custom Pre-processing}, which includes works that employ specific, non-standard data pipelines; and
ii) \textit{Benchmark Pre-processing}, which includes methods, such as ours, that follow the standardized and real-world plausible data preparation and splitting protocol proposed in \cite{nature-survey}. \rev{The two categories are discussed in Section~\ref{sec:custom-preproc} and \ref{sec:benchmark-preproc} respectively.}
\rev{Another limitation of most of the available works is that they do not analyze the performance of their algorithms once deployed on wearable-class hardware, thus making their real world usability without cloud support questionable. Some of the few related works that investigate this aspect are thus discussed in Section \ref{sec:hw}.}

\subsection{Custom Pre-processing}\label{sec:custom-preproc}

The authors of \cite{cuffless2017} addressed BP estimation using the publicly available MIMIC-II dataset \cite{mimic_ii}. After filtering out invalid records (e.g., short duration), they retained 3663 segments from approximately 1000 unique patients. Their pre-processing involved wavelet transformation and signal zeroing to mitigate noise and artifacts.
The processed data were evaluated using classical ML models, with AdaBoost achieving the best performance, yielding MAEs of 5.35 and 11.17 for DBP and SBP, respectively. In 2021, \cite{sensors_dataset} introduced a new dataset composed of recordings from 1195 Intensive Care Unit (ICU) patients. Their pre-processing involved filtering techniques and min-max normalization. The authors proposed a DL architecture consisting of an encoder, a decoder, and attention modules. The model achieved MAEs of 6.57 (DBP) and 14.39 (SBP). \cite{ppg-dnn-2} proposed a novel neural network architecture inspired by Wave-U-Net~\cite{stoller2018waveunetmultiscaleneuralnetwork}, which processes the PPG signal along with its first and second derivatives (VPPG and APPG, respectively). Using the MIMIC-II dataset~\cite{mimic_ii}, cleaned by removing short and unreliable signals, they obtained 277050 segments from approximately 1620 patients. The model achieved MAEs of 3.73 (DBP) and 6.41 (SBP).
In 2023, \cite{10147220} conducted a comprehensive comparison of classical algorithms (e.g., XGBoost, LightGBM, CatBoost) and deep learning models (e.g., Residual U-Net, ResNet-18, ResNet-LSTM) on both MIMIC-II and MIMIC-III datasets \cite{mimic_ii, mimic_iii}. Pre-processing varied by dataset: for MIMIC-II, a fourth-order band-pass filter was applied to the PPG signal, whereas for MIMIC-III, preprocessing involved removing ABP values smaller than 20 mmHg and higher than 300 mmHg, along with the same band-pass filter. Classical models outperformed deep learning methods, with XGBoost achieving the lowest MAEs on MIMIC-III (DBP and SBP), and on MIMIC-II (DBP), while CatBoost performed best on MIMIC-II (SBP). More recently, \cite{Tian2025} proposed the Parallel Convolution Transformer Network, which integrates convolutional and transformer layers to capture both local and global signal features. Evaluated on the MIMIC-III dataset, the model outperformed previous methods, achieving MAEs of 2.36 (DBP) and 4.44 (SBP), setting a new state-of-the-art on this dataset. However, this performance is conditioned by the fact that authors discard low-quality segments, therefore not being in the same testing condition of competitors.

\begin{table*}[ht]
\centering
\caption{State-of-the-art table of methods with Custom data pre-processing}
\begin{tabular}{|c|c|c|c|c|c|}
\hline
\textbf{Work} &
  \textbf{Dataset} &
  \textbf{Architecture} &
  \textbf{Preprocessing} &
  \textbf{DBP MAE} &
  \textbf{SBP MAE} \\
  \hline
  \noalign{\vskip 2pt} 
\noalign{\vskip 2pt} 
\hline

\multirow{4}{2cm}{\centering \textit{Kachuee et al.}, 2017 \cite{cuffless2017}} & \multirow{4}{*}{MIMIC II} & Linear Regression           & \multirow{4}{4.5cm}{\centering Re-sampling 1kHz, Wavelet decomposition, Zeroing [0, 0.25]Hz, Zeroing [250,500]Hz, Wavelet denoising, Wavelet reconstruction} & 6.74 & 14.71 \\
&                            & Decision Tree &             & 7.75 & 16.28 \\
&                            & SVR         &                                                              & 5.91 & 12.26 \\
&                            & AdaBoost      &                                                              & 5.35 & 11.17 \\
\hline\hline

\multirow{5}{2cm}{\centering \textit{Aguirre et al.}, 2021 \cite{sensors_dataset}}&
\multirow{5}{*}{Sensors} &
\multirow{5}{*}{\shortstack[c]{Encoder, Decoder \\ and Attention}} & \multirow{5}{4cm}{
\centering Null data and saturated points detection, Butterworth Filter [0.5, 8] Hz, Min-Max Normalization, band-pass Butterworth filter [0.5, 45] Hz} &
\multirow{5}{*}{6.57} &
\multirow{5}{*}{14.39} \\
& & & & & \\
& & & & & \\
& & & & & \\
& & & & & \\

\hline
\hline

\multirow{2}{2cm}{\centering \textit{Cheng et al.}, 2021 \cite{ppg-dnn-2}} & \multirow{2}{*}{MIMIC-II} & \multirow{2}{*}{ABP-Net} & \multirow{2}{4cm}{\centering Removal of short and unrealiable signals} & \multirow{2}{*}{3.73} & \multirow{2}{*}{6.41} \\
& & & & & \\

\hline
\hline

\multirow{12}{2cm}{\centering \textit{Costa et al.}, 2023 \cite{10147220}} & \multirow{6}{*}{MIMIC-II} & XGBoost & \multirow{6}{4cm}{\centering Cleaning and removal of dirty samples, PPG passed through a fourth-order Chebyshev II band-pass filter from 0.5 Hz to 10 Hz} & 7.14 & 16.67 \\
&  & LightGBM & & 7.21 & 16.49 \\
&  & CatBoost & & 7.38 & 16.43 \\
&  & Residual U-Net & & 8.18 & 19.1 \\
&  & ResNet 18 & & 8.42 & 19.85 \\
&  & ResNet LSTM & & 7.96 & 17.71 \\ \cline{2-6}
& \multirow{6}{*}{MIMIC-III} & XGBoost & \multirow{6}{4cm}
{\centering Cleaning of missing signals and setting of upper and lower bound for the ABP (300mmHg and 20mmHg)} & 8.45 & 16.18 \\
&  & LightGBM & & 8.59 & 16.28 \\
&  & CatBoost & & 8.60 & 16.32 \\
&  & Residual U-Net & & 10.94 & 18.60 \\
&  & ResNet 18 & & 10.38 & 18.16 \\
&  & ResNet LSTM & & 10.47 & 17.56 \\

\hline
\hline

\multirow{4}{2cm}{\centering \textit{Tian et al.}, 2025, \cite{Tian2025} } & \multirow{4}{*}{MIMIC-III} & \multirow{4}{2cm}{\centering Parallel Convolutional Transformer Network} & \multirow{4}{4.5cm}{\centering Cleaning unpaired PPG and ABP data, removal of data with poor signal quality, band-pass filter with frequency range of [0.5, 10]Hz} & \multirow{4}{*}{2.36$^*$} & \multirow{4}{*}{4.44$^*$} \\
& & & & & \\
& & & & & \\
& & & & & \\
\hline
\end{tabular}
\label{tab:related-works-others}
\begin{flushleft}
\footnotesize $^{*}$ Low-quality segments discarded and not tested
\end{flushleft}
\end{table*}

\begin{table*}[ht]
\centering
\caption{State-of-the-art table of methods with standardized pre-processing from \cite{nature-survey}. \rev{The bold numbers represents the most accurate models on the four datasets on both DBP and SBP.}}
\begin{tabular}{|c|c|c|c|c|c|}
\hline
\textbf{Work} &
  \textbf{Dataset} &
  \textbf{Architecture} &
  \textbf{Preprocessing} &
  \textbf{DBP MAE} &
  \textbf{SBP MAE} \\
  \hline
\noalign{\vskip 2pt} 
\noalign{\vskip 2pt} 

\hline
\multirow{8}{2.5cm}{\centering \textit{Gonzalez et al.}, 2023 (Benchmark) \cite{nature-survey}} & Sensors & \multirow{4}{*}{ResNet} & \multirow{8}{*}{\shortstack[c]{Standard benchmark preprocessing \\ (no fine-tuning)}} & 8.33 & 17.46 \\
& UCI &  &  & 8.30 & 16.59 \\
& BCG &  &  & 7.76 & 12.20 \\
& PPGBP &  &  & \textbf{8.61} & \textbf{13.62} \\
\cline{2-3} \cline{5-6}
& Sensors & \multirow{4}{*}{U-Net} & & 7.66 & 15.64 \\
& UCI &  &  & 7.88 & 16.93 \\
& BCG &  &  & 7.98 & 12.30 \\
& PPGBP &  &  & \textit{n/a} & \textit{n/a} \\

\hline
\hline

\multirow{2}{2cm}{\centering \textit{Lim et al.}, 2025 \cite{Lim2025}} & \multirow{2}{*}{PPGBP} & \multirow{2}{*}{Conv-Transformer} & \multirow{2}{4.5cm}{\centering Standard benchmark preprocessing (no fine-tuning)} & \multirow{2}{*}{9.17} & \multirow{2}{*}{14.82} \\
& & & & & \\
\hline
\hline

\multirow{8}{2.5cm}{\centering This Paper, 2025}
& Sensors & \multirow{4}{*}{ResNet} & \multirow{8}{*}{\shortstack[c]{Standard benchmark preprocessing \\ (no fine-tuning)}} & 7.83 & 17.01 \\
& UCI     &                         &                                                  & \textbf{7.69} & 17.09 \\
& BCG     &                         &                                                 & 7.50 & 11.51 \\
& PPGBP   &                         &                                                  & 8.68 & 13.88 \\ \cline{2-3} \cline{5-6}
& Sensors & \multirow{4}{*}{U-Net}  &  & \textbf{7.51} & \textbf{15.51} \\
& UCI     &                        &                                                  & 7.81 & \textbf{16.32} \\
& BCG     &                        &                                                  & \textbf{7.26} & \textbf{11.11} \\
& PPGBP   &                        &                                                  & \textit{n/a$^*$} & \textit{n/a$^*$} \\

\hline
\end{tabular}
\label{tab:related-works-benchmark}
\begin{flushleft}
\footnotesize \rev{$^{*}$U-Net, which is a signal-to-signal model cannot be trained on PPGBP because the dataset just contains the scalar values of SBP and DBP.}
\end{flushleft}
\end{table*}

However, due to the diversity in data processing and validation methods, comparing these results remains difficult. Moreover, many of these models are tested on only a single dataset, making their generalizability questionable.

\subsection{Standardized Pre-processing}\label{sec:benchmark-preproc}
To address this issue, in 2023, the authors of \cite{nature-survey} proposed a standardized preprocessing pipeline for four datasets (Sensors \cite{sensors_dataset}, UCI \cite{uci_dataset}, BCG \cite{BCG_dataset}, and PPGBP \cite{PPGB_dataset}). This pipeline significantly reduces result variability caused by data leakage and inconsistent pre
-processing. Their methodology includes:
\begin{itemize}
    \item Alignment of PPG and ABP signals using maximum cross-correlation
    \item Segmentation into non-overlapping 5-second windows
    \item Removal of distorted ABP segments based on physiological plausibility (e.g., amplitudes between 30-220 mmHg, pulse pressure over 10 mmHg, resting heart rate between 35-140 BPM)
    \item \rev{Extraction of SBP and DBP labels from ABP for signal-to-value models via the median of systolic peaks and cardiac cycle boundaries (refer to Figure \ref{fig:BP_PPG}, right, for a visual representation).}
    \item Removal of distorted PPG segments based on peak/valley standard deviation and baseline correction via cubic spline interpolation.
\end{itemize}

In Table \ref{tab:related-works-benchmark}, we report the results of the two best-performing DL models (on average) from \cite{nature-survey}, which are those that we used as input for our optimization pipeline (i.e., ResNet and U-Net).
In \cite{Lim2025}, similarly to \cite{Tian2025}, the authors introduced a hybrid architecture combining convolutional and transformer layers designed to extract both local and global features. They tested their model on PPGBP using the standardized benchmark pipeline. Despite the innovative architecture, their model underperformed compared to a standard ResNet of \cite{nature-survey}, achieving MAEs of 9.17 vs. 8.61 (DBP) and 14.82 vs. 13.62 (SBP), respectively.

To ensure fair and reproducible comparisons, our study adopts the standardized benchmark pre-processing introduced in~\cite{nature-survey}. This guarantees data leakage–free evaluation across subjects and aligns with best practices in the field, while providing a robust state-of-the-art comparison. \rev{In particular, rather than performing standard sample-level random splits (which could place measurements from the same subject in both training and test partitions), we enforce subject-wise partitioning so that all samples from a given subject are assigned to a single fold/set.}

\rev{Even though several works in Table \ref{tab:related-works-others} report promising performance (e.g., \cite{Tian2025}), we do not include them in our comparison for two reasons: (i) their custom pre-processing pipelines hinder a fair evaluation; in particular, the commonly adopted random train/test split can place samples from the same subject in both sets, leading to overly optimistic metrics; and (ii) they rely on operators that are too computationally demanding to be efficiently deployed on wearables (e.g., Transformers \cite{attention-is-all-you-need}).  In contrast, our pipeline (architecture
exploration, structured pruning, and integer-only quantization) systematically
optimizes edge-friendly CNNs using a robust and leakage-free data pipeline. Within that experimental setup,  it not only results in models with a smaller memory footprint with respect to previous DNN-based solutions, but also superior accuracy.
}

\subsection{Embedded solutions}
\label{sec:hw}
\rev{
While pre-processing  and data splitting plays a crucial role in ensuring a fair comparison across methods and measuring accuracy realistically, an equally important requirement for real-world applicability is deployment on wearable device-class hardware.}

\rev{
In \cite{Zhang2026}, the authors provide a comprehensive analysis and comparison of approaches for embedded BP estimation. They review a broad range of deployment platforms, spanning commercial processors (CPUs and GPUs), general-purpose microcontrollers, and custom hardware solutions such as FPGAs \cite{10798356} and ASICs \cite{10.1145/3699512, 9871215}. Although FPGA- and ASIC-based implementations can be extremely efficient, we do not consider them in this study because they are arguably less applicable to real-world products (unless with huge market volumes), due to their high design costs, especially for ASICs, and limited flexibility. In fact, those approaches usually implement a \textit{fixed algorithm} or DNN structure, and require an external host for control, reconfiguration (if any), sensor data reading and transmission, etc. In contrast, we focus our analysis on solutions deployed on \textit{programmable} embedded devices, describing below some of the most notable previous works. While we also report their MAE results, besides deployment metrics, we remark that none of them uses the same data splitting and pre-processing setup of~\cite{nature-survey}, thus error values are not directly comparable with ours.
}

\rev{In \cite{9630488}, the authors propose a simple two-layer ANN that takes the raw PPG waveform as input and directly estimates blood pressure. The model is deployed on an ultra-low-power EFM32 Leopard Gecko microcontroller (ARM Cortex-M3, 32-bit). While the reported energy per inference is very low (about 2.1 mJ), the inference latency is still relatively high (about 42 ms) considering the simplicity of the network. The memory consumption is reported to be less than 25kB. On MIMIC-III, they obtained a MAE of 3.42 and 1.92 for SBP and DBP respectively.
In \cite{9680071}, the authors introduce BP-Net, a neural architecture derived from U-Net \cite{unet}. They deploy the model on a Raspberry Pi 4 with 4 GB of RAM and report an inference time of 42.5 ms, comparable to \cite{9630488} despite the substantially more complex architecture and much more capable hardware. Although memory and power are not reported, a conservative upper bound for the latter can be derived from the platform supply rating ($\approx$15 W), which would result in a very high energy consumption per inference of $\approx$ 0.64 J.
The new architecture was evaluated on MIMIC-II obtaining a MAE of 5.16 and 2.89 on SBP and DBP respectively.
In \cite{10.1145/3555776.3577747}, the authors evaluate several deep networks (including a 0.30MB ResNet) across multiple edge devices, such as the Raspberry Pi 3 Model B and the ESP32-WROVER-IE. For ResNet on the Raspberry Pi, they report an inference time of 186 ms. Regarding energy, the same considerations done for the previous paper apply (power values are not reported either). On a more energy-constrained platform, the ESP32, the inference time increases dramatically to 60.82 s, which effectively rules out real-time use. Evaluated on the MIMIC-IV dataset, their ResNet20 achieved a MAE of around 12.1 and around 5.9 for SBP and DBP respectively. 
In \cite{9913482}, the authors deploy a Temporal Convolutional Network (TCN) that uses PPG features as input on a Raspberry Pi Zero W. They report an inference time of about 2.5s for a 3s PPG window. Power and energy metrics are again not reported, while memory is reported to be around 32kB. On MIMIC-II/III the authors achieved a MAE of 2.38 (SBP) and 1.23 (DBP).
Finally, in \cite{10856555}, the authors deploy a substantially more complex model (SCI-GTCN, Sample Convolution and Interaction Network and Gated Time Convolutional Network) on an NVIDIA Jetson Nano. The inference time is around 2s and no memory consumption is reported; although power is also not reported, the Jetson Nano is far more power-hungry than the ultra-low-power embedded targets considered in this work. On the MIMIC-III dataset, they obtained a MAE of 7.44 (SBP) and 5.70 (DBP). 
}

\section{Materials \& Methods}

\rev{The main goal of this paper is to define a complete, multi-stage DNN optimization pipeline for PPG-based BP estimation, targeted at obtaining models that are not only accurate, but also compact, thus amenable for deployment on wearables. To this end, we extended the open-source DNN optimization library PLiNIO~\cite{plinio}, which offers a user-friendly interface for implementing a variety of model selection and compression algorithms. While the individual algorithms used in this work were already included in the PLiNIO package, in this work we: i) integrated them into a coherent pipeline and ii) extended them to be applicable onto state-of-the-art DNN architectures for BP estimation, such as U-Net-like autoencoders. This required multiple modifications to the PLiNIO inner workings and the addition of support for new layer types.}

\begin{figure}[ht]
    \centering
    \includegraphics[width=.8\columnwidth]{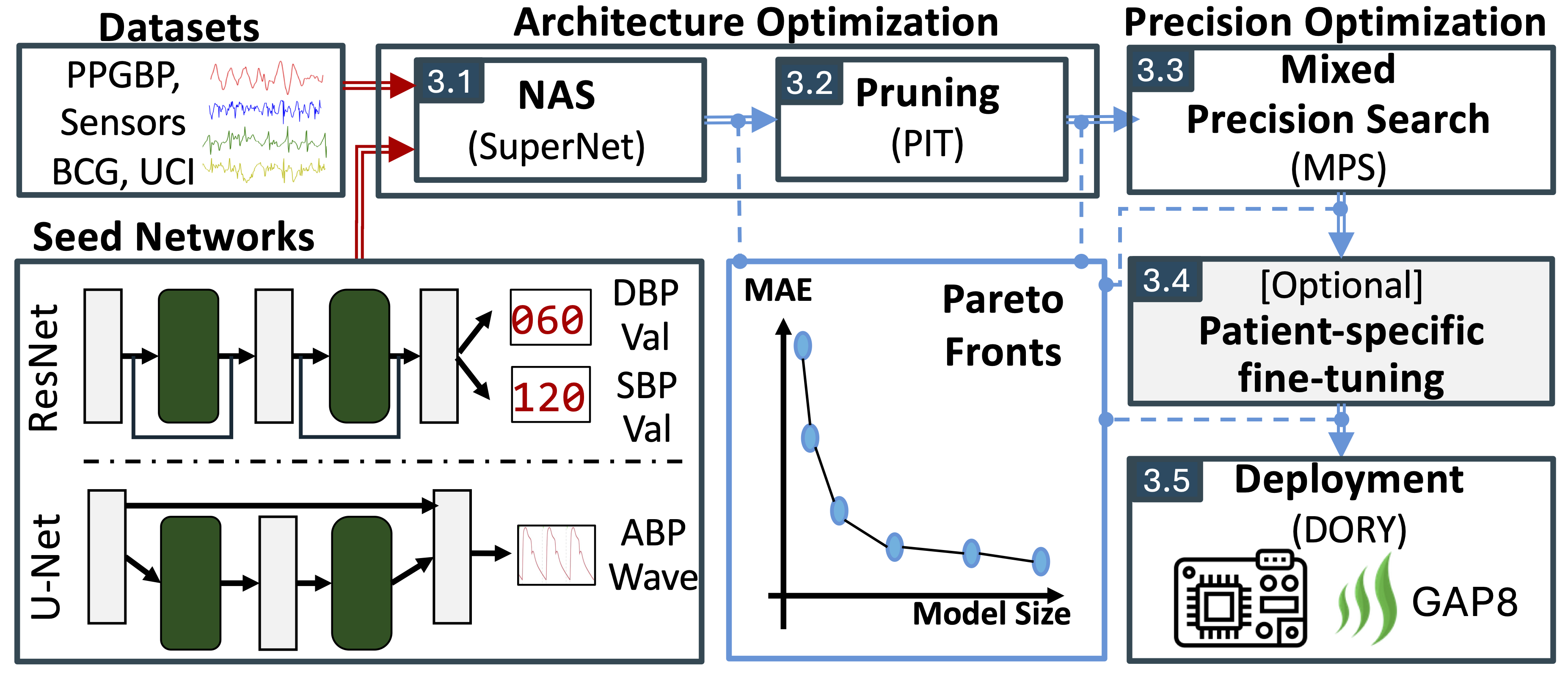}
    \caption{\rev{Overview of the proposed automated DNN optimization flow. Starting from a dataset and a baseline model, the pipeline performs Neural Architecture Search (NAS), followed by structured pruning and mixed-precision search (quantization). Before deployment, selected candidate models can optionally undergo patient-specific fine-tuning using data from an individual subject to further improve accuracy.}}
    \label{fig:flow}
\end{figure}
Our pipeline, depicted in Figure \ref{fig:flow}, is composed of three model optimization steps: Neural Architecture Search (\textit{SuperNet}), structured pruning (\textit{Pruning-In-Time}), and Mixed Precision Search (\textit{MPS}). We (optionally) apply a subject-specific fine-tuning step to further improve the accuracy of our optimized models, followed by an automatic Python-to-C compilation and deployment on the target SoC.

The optimization process takes as input a training dataset, preprocessed and segmented into windows as described in~\cite{nature-survey}, alongside a \textit{seed} network. This seed, representing the initial DNN architecture, acts as a template from which optimized models are derived. We adopt as seeds the two top-performing DNNs identified in~\cite{nature-survey}. Both are 1D CNNs; however, they differ in their design and target output. The first is a signal-to-value model inspired by ResNet~\cite{resnet-ppg}, which directly predicts the average SBP and DBP values within each input window.
The second model follows a U-Net-like~\cite{unet} signal-to-signal structure, estimating the entire ABP waveform from which SBP and DBP values are later extracted. For more detailed information on these seed architectures, we refer the reader to~\cite{nature-survey}.

Importantly, while these models were previously optimized in~\cite{nature-survey} for maximum predictive accuracy, our work introduces a new dimension: \textit{hardware cost-aware} optimization. We demonstrate that this approach enables the discovery of models with significantly reduced size and computational demands.
For each combination of seed network and dataset, our optimization pipeline follows three core stages, described in the next sections.

\subsection{SuperNet NAS}\label{sub:supernet}

\begin{figure}[ht]
    \centering
    \includegraphics[width=0.7\columnwidth]{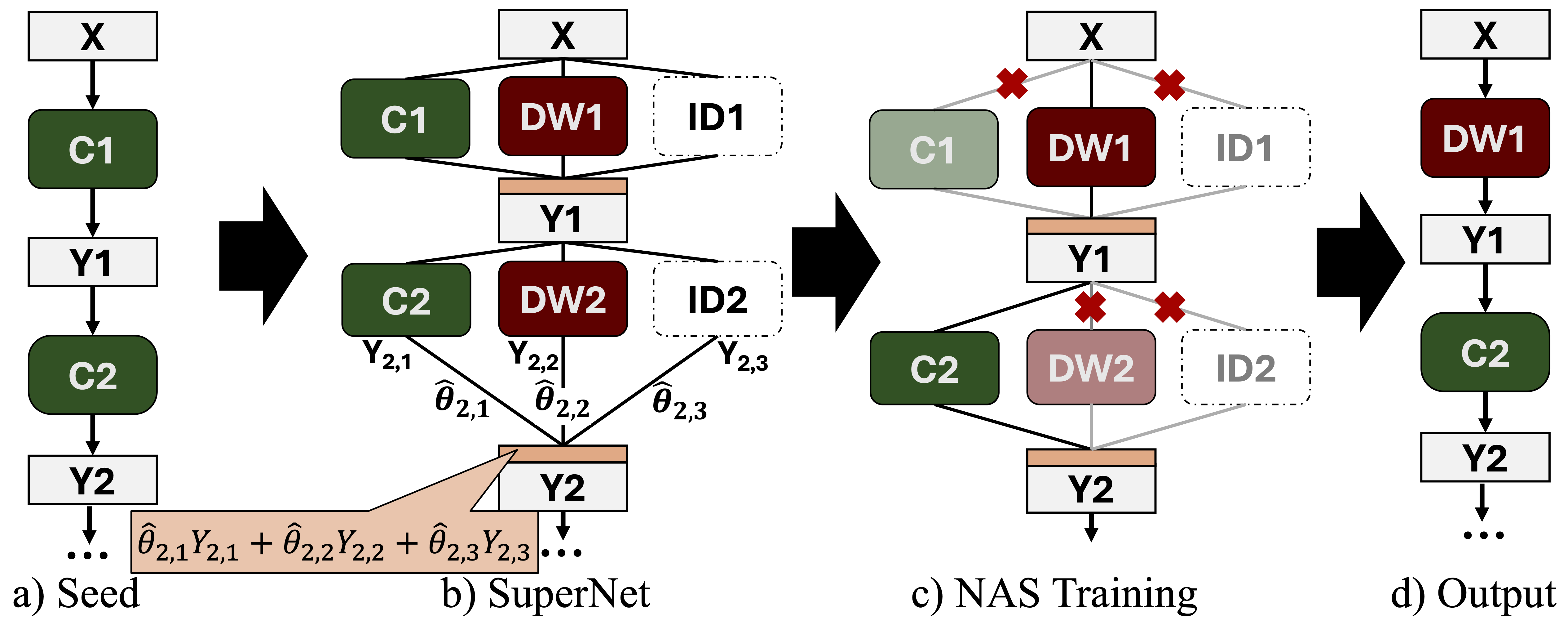}
    \caption{\rev{SuperNet-based NAS. From left to right, the figure reports the initial condition of the network (a), the different layer options (b), the selection process during training (c) and finally the output model (d).}}
    \label{fig:supernet}
\end{figure}

The first step of our pipeline consists of applying a gradient-based NAS, whose working principle, shown in Figure \ref{fig:supernet}, is inspired by \rev{DARTS} \cite{darts}. \rev{Compared to the latter, SuperNet supports hardware-aware optimization, considering various cost models (e.g., model size, number of operations) and is able to generate multiple architectures, thereby exploring the accuracy-cost trade-off more thoroughly.}
The NAS builds a so-called SuperNet, replacing every convolutional layer $l_i$ in the seed architecture with a \textbf{set \rev{$S_i$} of alternative layers $l_{i,j}$}, all of which receive the same input.
\rev{Here, $i$ indexes the position of the original layer in the seed network (i.e., the $i$-th convolution being replaced), while $j$ indexes a specific candidate operation among the alternatives available for that layer (i.e., the $j$-th option in the set $S_i$).}
In our work, each standard convolution from the two seeds is replaced with a set of alternatives \rev{chosen specifically to reduce the models' footprint towards wearable deployment}, including: a standard 1D convolution (C), an identity layer (ID), and a depthwise-separable convolutional module (DW). The latter is composed of a depthwise layer followed by a pointwise convolution, and it is commonly used in efficient CNN models due to its reduced parameters and FLOPs count w.r.t. standard convolution \cite{mobilenet}.
The Identity operation, instead, is added only when the input and output of the original layer have the same shape; it allows the NAS to skip some convolutions, thus optimizing the network depth.
Each alternative $l_{i,j}$ in the set has a corresponding trainable parameter $\theta_{i,j}$. The output of the set is computed as a weighted sum of the various alternatives’ outputs, where the weight associated with each alternative $l_{i,j}$ depends on a softmax-ed version of the corresponding $\theta_{i,j}$ as depicted in Figure \ref{fig:supernet}b.

Formally, let \(S_i = \{l_{i,1}, l_{i,2}, ..., l_{i,n}\} \) be the set of alternative paths replacing a single original layer. Given the layer's input \( X_i \), the output \( Y_i \) of the resulting SuperNet layer is computed as shown in Eq.\ref{eq:weighted_output}:

\begin{equation}
Y_i = \sum_{j = 1}^{||S_i||} \hat{\theta}_{i,j} \cdot Y_{i,j}
    \qquad \text{with \ $\hat{\theta}_{i,j} = \sigma_j(\theta_{i,j})$ }
\label{eq:weighted_output}
\end{equation}

where $\sigma()$ is the softmax operation and \( Y_{i,j} = l_{i,j}(X_i) \) is the output of layer \( l_{i,j} \).

A NAS-optimized architecture is obtained, intuitively, by choosing for each layer a single path, i.e., setting one of the parameters $\hat{\theta}_i$ to $1$ and the other to $0$.
\rev{SuperNet uses gradient descent to solve a continuous relaxation of this assignment problem, jointly optimizing both the weights $W$ and the added $\theta$s in a standard training loop.} 

This training uses the modified loss function shown in Eq. \ref{eq:total_loss}, where the standard task loss $\mathcal{L}$, computed on the model's prediction and on the ground truth BP values is augmented by a cost-based regularization term $\mathcal{R}$. In our case $\mathcal{L}$ is the Mean Squared Error (MSE) of the SBP and DBP predictions for ResNet, or of the whole reconstructed signal for U-Net.

\begin{equation}
\mathcal{L}(W, \theta) = \mathcal{L}(W, \theta) + \lambda \mathcal{R}(\theta)
\label{eq:total_loss}
\end{equation}

\( \mathcal{R}\), defined in Eq. \ref{eq:cost_term}, estimates the expected cost of the architecture by weighting the cost of each alternative layer.

\begin{equation}
\mathcal{R}(\theta) = \sum_{i}\sum_{j = 1}^{||S_i||} \hat{\theta}_{i,j} \cdot Cost(l_{i,j})
\label{eq:cost_term}
\end{equation}
$\text{Cost}(l_{i,j})$ represents a predefined cost metric (e.g., the number of parameters associated with each alternative), and $\lambda$ is a hyperparameter that balances accuracy and efficiency.
In our setup, we use the parameter count as a cost metric to favor architectures suitable for deployment on memory-constrained wearable platforms.
When the training is finished, the optimized architecture is obtained keeping the path associated with the largest $\theta_i$, discarding the others. 

\rev{Algorithm \ref{pseudocode_supernet} reports a high level overview of the Differential NAS (DNAS) training procedure. 
 It is composed of three phases: the SuperNet is initially pre-trained with all paths equally contributing to the output; then, the actual DNAS optimization is started; lastly, the model can be fine-tuned after selecting the best single path.
 This general scheme also applies to the following optimizations (pruning and mixed precision quantization), which similarly employ a gradient-based approach.
However, in later pipeline steps some of the phases may become redundant and can be skipped. For instance, an already fine-tuned model coming from the SuperNet optimization does not require additional pre-training (phase 1), and can be directly pruned
(phase 2). The same happen when applying mixed precision search to an already optimized network.
For further details about the DNAS training procedure and all other algorithms employed in this work we refer the reader to \cite{plinio}.
}

\begin{algorithm}[t]
\caption{Training-time optimization procedure}
\label{pseudocode_supernet}
\begin{algorithmic}[1]
\FOR{\textbf{each} $l_i \in \text{SuperNet}$}
    \FOR{$l_{i,j} \in S_i$}
        \STATE Add $\theta_{i,j} \gets 1 / ||S_i||$ \textcolor{olive}{$\triangleright$ Initialization}
    \ENDFOR
\ENDFOR

Freeze all $\theta$
\FOR{ $i \gets 1, \ldots, \text{Epochs}_{pt}$}
    \STATE Update $W$ based on $\mathcal{L}(W)$ \textcolor{olive}{$\triangleright$ Pre-training (phase 1)}
\ENDFOR

Unfreeze all $\theta$
\WHILE{not converged}
\STATE Update $W,\theta$ based on $\mathcal{L}(W, \theta) + \lambda \mathcal{R}(\theta)$ \textcolor{olive}{$\triangleright$ DNAS (phase 2)}
\ENDWHILE

Extract selected path based on $\hat{\theta}$
\FOR{ $i \gets 1, \ldots, \text{Epochs}_{ft}$}
    \STATE Update $W$ based on $\mathcal{L}(W)$ \textcolor{olive}{$\triangleright$ Fine-tuning (phase 3)}
\ENDFOR

\end{algorithmic}
\end{algorithm}

\subsection{Structured Pruning}

\begin{figure}
    \centering
    \includegraphics[width=0.7\columnwidth]{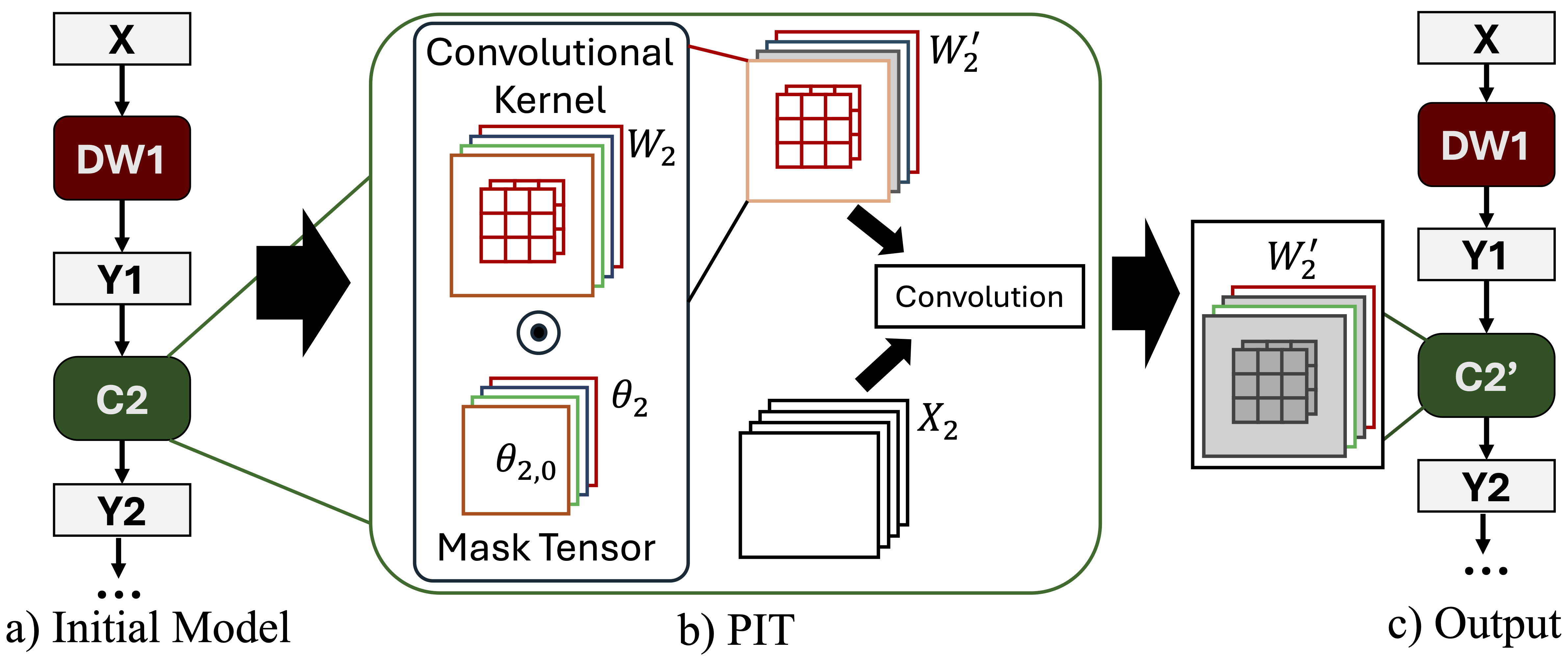}
    \caption{\rev{Pruning-in-Time (PIT) overview: starting from the initial layer sequence (a), PIT injects a trainable mask tensor \( \theta_2 \) to gate slices of the convolutional kernel \(W_2\) during training (b), yielding a pruned kernel \(W_2'\) and a reduced layer \(C2'\) in the final network (c).}}
    \label{fig:pit}
\end{figure}

Starting from the optimized network architectures found by the NAS, the second optimization step consists in applying structured pruning, to further reduce their parameter count (and indirectly, their computational complexity) while preserving accuracy.

We employed the Pruning-in-Time (PIT) method from PLiNIO, which prunes convolutional and linear layers by removing slices of the weight tensors (and corresponding activations) over multiple axes, with the effect of: i) reducing the number of output channels/neurons; ii) reducing the filter size (i.e., the receptive field) or iii) increasing the dilation.
Thus, this step implements a finer-grain neural network optimization compared to the \textit{NAS}, tuning each layer’s geometrical hyper-parameters rather than selecting among pre-defined alternatives, possibly improving the Pareto front found in the previous step.
\textit{PIT} works by adding trainable binary masks to a model, that selectively eliminate slices of weights and activation tensors. These masks are once again optimized jointly with the network weights via gradient descent, using the loss formulation of Eq.~\ref{eq:total_loss}. After training, pruned parts are permanently removed. Figure \ref{fig:pit} shows an example of the pruning method when applied to \textit{output channels}. As shown, a trainable vector of binary mask parameters $\theta_i$ is added to each layer, where each $\theta_{i,j}$ determines whether an output channel $j$ is kept (setting $\theta_{i,j}=1$) or removed ($\theta_{i,j}=0)$. The masks are applied to the weights $W_i$ of layer $l_i$ as follows:
\[
W_i' = H(\theta_i) \circ W_i
\]
where $H(\theta_i)$ is the Heaviside step function and $\circ$ denotes the Hadamard product. Since the Heaviside function is non-differentiable, the \textit{Straight-Through Estimator} (STE) is used to allow gradients to flow through $\theta_i$ during backpropagation.

\rev{As anticipated in Section~\ref{sub:supernet}, PIT's optimization is carried out following a scheme similar to the one of Algorithm~\ref{pseudocode_supernet}, albeit without the initial pre-training phase. Moreover, PLiNIO's original PIT algorithm had to be extended to support state-of-the-art BP prediction networks. The main changes were required to add support for pruning some previously unsupported layer types, such as Parametric ReLUs (PReLUs), Instance Normalization, features concatenation, and grouped convolutions. In the latter, pruning output channels independently as done normally by PIT would lead to an irregular result, with different groups including a different number of filters. Such a layer would be harder to accelerate, and incompatible with standard DNN frameworks (e.g. PyTorch). Therefore, we added an extra preparation step to ensure that \textit{channels are pruned uniformly across groups}, by sharing the masks relative to corresponding channels within different groups (e.g., the first output channels generated by group 1, group 2, etc, all share a single $\theta_i$).}

\subsection{Quantization and Mixed-Precision Search} 

After the network architectures have been optimized through NAS and further refined by PIT, the final stage before deployment is integer \textit{quantization}. This step further reduces the memory footprint and computational requirements of the models by representing weights and activations as low-precision integer values, while also enabling efficient deployment on hardware platforms that do not include a Floating Point Unit (FPU), such as our target, GAP8~\cite{gap8}, and many other low-power wearable-class devices. PLiNIO supports Mixed-Precision Search (MPS), which assigns independent precision levels to weights ($W$) and activations ($X$) in Convolutional and Fully Connected layers. Inspired by \cite{cai2020rethinkingdifferentiablesearchmixedprecision}, this method can be configured to select from a set of predefined bit-widths $p \in P$, e.g., $P = \{2, 4, 8\}$.

\begin{figure}[ht]
    \centering
    \includegraphics[width=0.7\columnwidth]{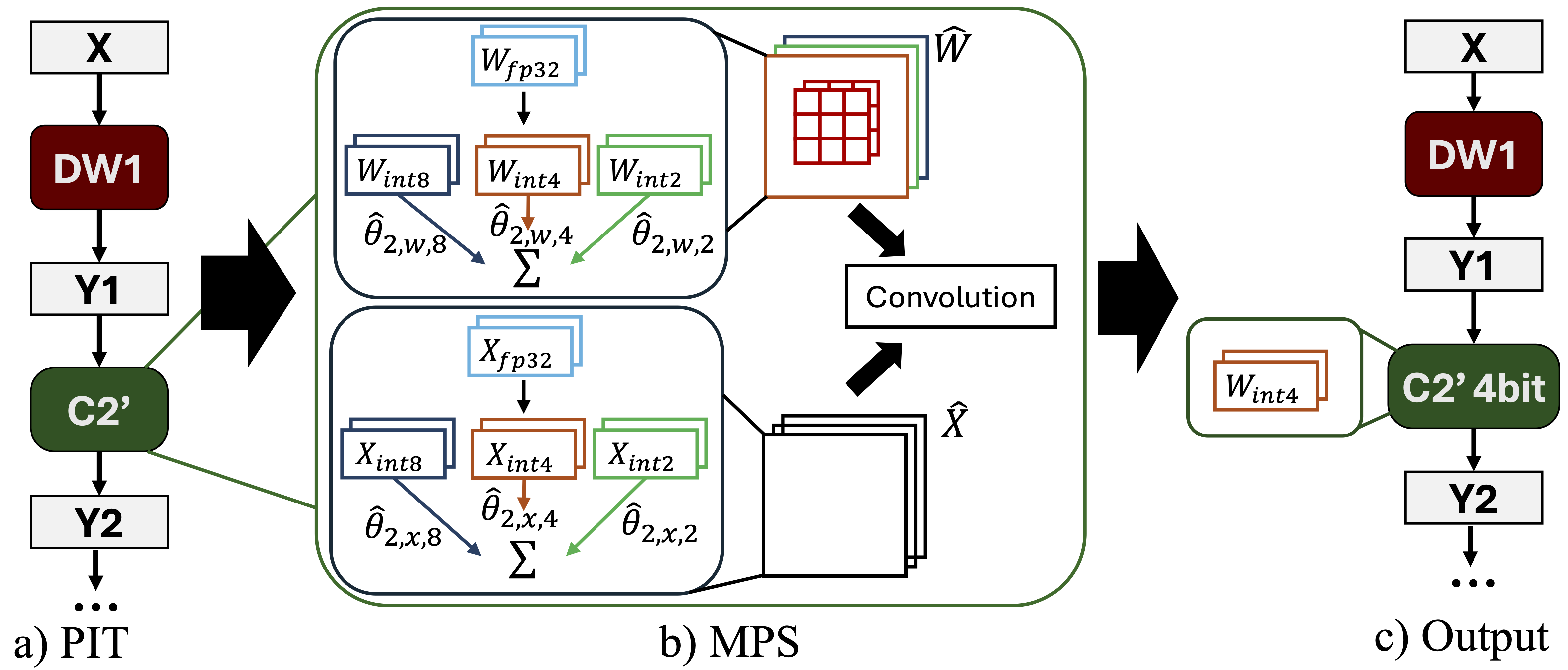}
    \caption{\rev{Mixed Precision Search (MPS). The figure shows the selection of the bitwidth of each convolutional layer, which leverages a SuperNet-like approach. Supported bitwidth are currently 2, 4, and 8.}}
    \label{fig:mps}
\end{figure}

MPS uses so-called ``fake-quantization'' during the training/optimization phase, i.e.,  it uses floating point values internally to allow small gradient-based updates, but simulates an affine quantization process by means of scaling, offsets, and rounding \cite{plinio}. The non-differentiable rounding function is approximated by means of STE during back-propagation. More specifically, as shown in Fig. \ref{fig:mps}, the activations $X_i$ and the weights $W_i$ tensors of each layer, undergo fake-quantization for all bit-widths in the set $P$. Then, similarly to the various layer alternatives in \textit{SuperNet}, the different quantized variants are linearly combined using trainable parameter vectors $\theta_{i,w}$ (for weights) and $\theta_{i,x}$, which have a length of $\|P\|$ and are normalized via a SoftMax function ($\sigma$).
The resulting effective tensor is computed as:

\begin{equation}
    \hat{T_i} = \sum_{p \in P} \hat{\theta}_{i,t,p} \cdot T_{i,p} 
    \qquad \text{with} \  \hat{\theta}_{i,t,p} = \sigma_p(\theta_{i,t,p})
    \label{eq:1}
\end{equation}

where $T$ (and subscript $t$) are used to refer to a generic tensor which could be $W$ (subscript $w$) or $X$ (subscript $x$) and $T_{i,p}$ is the $p$-bit quantized version of $T_i$. A higher value of $\theta_{i,t,p}$ increases the contribution of the corresponding bit-width, making $\hat{T_i}$ more closely resemble the $p$-bit quantized tensor.
The effective tensors $\hat{W}$ and $\hat{X}$ derived in this way are then used to compute the layer’s output, e.g.:

\begin{equation}
    Y = \text{Conv}(\hat{X}, \hat{W})
    \label{eq:2}
\end{equation}

A key advantage of this approach is that all fake-quantized versions originate from a single floating-point tensor, significantly reducing memory overhead during training.

As with other PLiNIO methods, this modified DNN is trained in a DNAS-like optimization loop, where both $W$ and $\theta$ are updated jointly to minimize Eq.~\ref{eq:total_loss}.

In this work, precision search has been applied only to weights, fixing activations to 8 bits, because our main objective was to compress the model weights so that they could fit the target platform (GAP8) onboard memory. Therefore, the optimization cost term ($\mathcal{R}$) in Eq.~\ref{eq:total_loss} in this case is set to the total number of \textit{bits} required by the network, as a function of each $\theta_{i,w}$.

\subsection{Subject-specific fine-tuning}\label{sec:fine_tuning}

Several previous studies have shown that, due to physiological differences between subjects, ML models trained on biosignals for BP estimation may perform better when exposed to data from the test subject during training \cite{ppg-dnn-2, 10147220}. For this reason, it is common practice among commercial devices, especially those used for medical purposes to perform a calibration or fine-tuning operation tailored for each subject, e.g., using data from a sphygmomanometer as ground truth, which may be repeated periodically \cite{Alexandre2023}. 

\rev{To this end, subject-specific fine-tuning has been integrated as the final pipeline stage to specifically address the challenges inherent to PPG-based blood pressure estimation. 
Specifically, we propose a novel fine-tuning procedure that follows strict constraints to avoid data leakages.
By accommodating individual cardiovascular characteristics, this fine-tuning procedure enables the resulting models to satisfy the stringent accuracy thresholds mandated for clinical-grade diagnostic devices.}

Namely, let a dataset consist of $N$ patients, each denoted by $p_i \in \mathcal{P}$, where $i = 1, \ldots, N$. Each patient $p_i$ has $M_{p_{i}}$ samples.  In all our experiments, we initially train our models using a 5-fold cross-validation strategy, consistent with the methodology used in \cite{nature-survey}, where the splitting unit is the patient. In each fold, we generate a training set $\mathcal{P}_{\text{train}}$ and a test set $\mathcal{P}_{\text{test}}$, such that \( \mathcal{P}_{\text{train}} \cup \mathcal{P}_{\text{test}} = \mathcal{P} \) and \( \mathcal{P}_{\text{train}} \cap \mathcal{P}_{\text{test}} = \emptyset \). Thus, models at this stage have not been trained on any sample from the subject under test. 

When performing experiments that require a direct comparison with solutions trained following the protocol of~\cite{nature-survey}, we simply evaluate trained models on all $M_{p_i}$ samples from $\mathcal{P}_{\text{test}}$. Specifically, \( |\mathcal{P}_{\text{train}}| = \frac{4N}{5} \) and \( |\mathcal{P}_{\text{test}}| = \frac{N}{5} \), supposing a 5-fold cross-validation. When assessing the benefits of subject-specific fine-tuning, the model is first evaluated on a subset of 20\% of the test subjects' samples to establish baseline performance. The remaining samples are used to fine-tune the model. Then, a final evaluation is again conducted on the same 20\% subset, allowing us to measure the improvement attributable to the fine-tuning step. Table \ref{tab:patient-specific-ft} summarizes the steps of the fine-tuning protocol.

\rev{
It is worth noting that our optimization pipeline (i.e. NAS, Pruning, Quantization) is applied to the state-of-the-art ResNet and U-Net backbones with the goal of obtaining more accurate and lightweight models that generalize across every subject, i.e., without requiring calibration. For this reason, the architectural parameters \( \theta \) are selected on a validation set that is strictly disjoint from the subjects in \( \mathcal{P}_{\text{test}} \), ensuring that model selection remains free of subject-specific leakage. Subject-specific fine-tuning is then studied separately, on a per-subject basis, starting from the model instances that achieve the best performance under the subject-independent evaluation protocol.
This paradigm ensures that the generated optimized architectures possess sufficient flexibility for subject specific fine-tuning on unseen data after they reached good performances across patients, thereby avoiding the risk of overfitting scarce data from individual subjects.
In real-world scenarios, our pipeline enables BP monitoring device providers to maintain a single high quality and efficient model that can be fine-tuned on limited user-specific data without requiring individual alterations to the architecture nor the deployed model code.
}
\begin{table}[ht]
\centering
\caption{Subject-specific fine-tuning protocol}
\label{tab:patient-specific-ft}
\renewcommand{\arraystretch}{1.5}
\begin{tabular}{|c|c|}
\hline
\textbf{Step} & \textbf{Description} \\
\hline
1 & Train the model on all samples from patients in \( \mathcal{P}_{\text{train}} \). \\
\hline
2 & Evaluate on 20\% of \( \mathcal{P}_{\text{test}} \) to obtain baseline accuracy. \\
\hline
3 & Fine-tune the model on the remaining \rev{samples of} \( \mathcal{P}_{\text{test}} \). \\
\hline
4 & Evaluate again on 20\% of \( \mathcal{P}_{\text{test}} \) to assess the effect of fine-tuning. \\
\hline
\end{tabular}
\end{table}

\subsection{Hardware Target and Deployment}
We export our optimized DNNs into a custom ONNX-compatible format, which can be directly processed by the open-source deployment framework DORY \cite{dory} for execution on multi-core RISC-V platforms. 
DORY automatically generates highly optimized C inference code, taking care of memory allocation, Direct Memory Access (DMA) transfer scheduling, and the invocation of optimized AI kernels. 
To maximize efficiency on our target platform, DORY applies graph-level optimizations, tiling strategies, and memory management techniques that exploit the hardware hierarchy, with the goal of minimizing latency and maximizing resource utilization.

For the implementation of DNN layers, we employ two distinct backend libraries of manually optimized DNN primitives in C for our target: PULP-NN \cite{pulp-nn} for fully \texttt{int8} quantized networks (a specific case explored in our MPS search), and 
PULP-NN-Mixed \cite{pulp-nn-mixed} for mixed-precision execution. 
The reason why we consider the full \texttt{int8} case separately is that this precision is natively supported by the 4-way, Multiply and Accumulate (MAC)-capable Single Instruction Multiple Data (SIMD) Arithmetic Logic Units (ALU) of our hardware target. Sub-byte operations, instead, require extra unpacking/packing operations onto/from 8-bit values \cite{pulp-nn-mixed}. 

Therefore, while MPS models including $<$ 8-bit tensors are surely beneficial in terms of memory compression, latency- and energy-wise their benefit is less certain. Indeed, smaller tensor sizes reduce the cost of memory transfer operations and allow fitting bigger portions of a layer in faster and closer memories, but this advantage can be reduced or nullified by the (un)packing overheads. Accordingly, we consider both the case in which MPS is allowed to select any precision in $P = \{2, 4, 8\}$ and the one in which it is constrained to $P=\{8\}$, thus becoming effectively equivalent to a standard Quantization-Aware Training (QAT).

\subsubsection{Hardware Platform and Measurements}
Our deployment target is the GreenWaves GAP8 SoC~\cite{gap8}, a low-power RISC-V-based multi-core processor specifically designed for edge signal processing. 
GAP8 features a cluster of eight general-purpose cores for parallel execution of compute-intensive workloads, a two-level scratchpad memory with 512\,kB of main memory for code and weight storage, and a 64\,kB last-level cache with single-clock access latency for the cluster. Data transfers between memory levels are managed by an integrated DMA engine. 

We use the GAPuino evaluation board as the deployment platform, connected to the Nordic Power Profiler Kit II~\cite{nordic-2} for accurate power and energy measurements. 
Execution latency is measured through the GAP8 internal performance counters, while the external profiler provides fine-grained insight into the energy consumption of the deployed models.

\rev{We note that, while GAP8 is a highly competitive ultra-low-power processor, featuring dedicated hardware acceleration capabilities for parallel, data-intensive workloads such as neural networks inference, the methodology described in the previous sections is not tied to this specific target. The generated models are entirely architecture agnostic, and could be easily deployed on alternative embedded architectures, albeit clearly with different results in terms of latency and energy consumption.}

\section{Results}
\subsection{Setup}

\rev{Some architectural modifications were required to ensure compatibility of our seed DNNs with \textit{full-integer} quantization and deployment, necessary to obtain good latency and energy performance on our target: Batch and Instance Normalization layers were repositioned immediately after convolutional layers, enabling their parameters to be folded into convolutional weights; similarly, PReLU activations were kept for SuperNet and PIT, but then replaced with standard ReLUs for the quantization step, to ensure compatibility with DORY~\cite{dory}. Additionally, zero-padding shortcuts in ResNet-based seeds were replaced with learnable $1 \times 1$ convolutions with independent normalization. This increases the number of parameters of the seed but also allows PIT to explore a larger optimization space.}

For the SuperNet NAS, we use two Adam optimizers: one for the network weights $\mathcal{W}$ (learning rate $0.001$) and one for the architecture parameters $\theta$ (learning rate $0.01$). Network weights were optimized on the training set, while $\theta$ on the validation set, following the DARTS approach~\cite{darts}. For each dataset and for both seeds, we explored 18 values of the regularization parameter $\lambda$, logarithmically spaced between $10^{-11}$ and $10^{-7}$, using the total parameter count as the optimization cost $\mathcal{R}$. NAS training consisted of $20$ warm-up epochs, followed by $200$ epochs of architecture search and an additional $200$ epochs of fine-tuning. Early stopping with a patience of $40$ epochs and checkpointing based on the minimum MSE loss were applied throughout. 

\rev{From the NAS results, we selected for PIT-based structured pruning the models achieving the lowest SBP and DBP errors, as well as the corresponding seed models even when they did not attain the lowest SBP/DBP errors.}
PIT training followed the same protocol as SuperNet with $200$ epochs of pruning optimization, and $200$ epochs of fine-tuning with early stopping. We also explored the same regularization strength values $\lambda$ as for the NAS.

In the final stage, we applied QAT (at fixed \texttt{int8} precision for both weights and activations) and MPS (assigning layer-wise weights bitwidths between \texttt{int2}, \texttt{int4}, and \texttt{int8}). \rev{As for the previous stage, we took as input both the seeds and the models with the lowest errors from the combined SuperNet and PIT Pareto fronts.} \rev{Additionally, only for the BCG dataset, we also applied QAT/MPS to the Pareto-optimal model with fewest parameters. The reason for this difference is that this is the dataset on which we measure deployment results (see Section~\ref{sec:deployment}), thus we wanted to also include an example of a much smaller (yet slightly less accurate) model.}
We applied a standard min–max affine quantization scheme for weights and signed Parametrized Clipping Activation (PaCT)~\cite{pact} for activations, with $32$-bit accumulations and biases, as supported by our target inference library~\cite{pulp-nn}. 
MPS training employed the same protocol as the previous stages: a differentiable search phase where $\mathcal{W}$ and $\theta$ were optimized jointly (with $\hat{\theta}$ annealed via a SoftMax temperature $\tau$ initialized at $5$ and decayed by $e^{-0.0045}$ per epoch \cite{10644100}), and a final fine-tuning phase where $\theta$ was frozen and discrete precision selections were applied. \rev{The parameter $\tau$ controls the smoothness of the sampling distribution. The annealing schedule is designed so that the bit-width selection parameters sampling increasingly resemble an argmax operation. Such procedure improves training stability and prevents degenerative conditions in which the $\theta$ parameters assume uniform values.}
We performed nine experiments per seed with logarithmically spaced values of $\lambda$ to balance the trade-off between model size and MSE loss.
\rev{Strong values of $\lambda$ can bring to aggressively quantized networks (i.e., employing many 2-bit and 4-bit weights). The output of such networks is a noisy reconstructed signal which can hinder the peak localization algorithm necessary to estimate BP values. To mitigate this problem}, a \rev{smoothing} filter was added to the final layer before extracting systolic and diastolic blood pressure values from reconstructed arterial waveforms. \rev{This fifth-order Butterworth low-pass filter employs an adaptive cutoff frequency defined as a proportional coefficent of the absolute magnitude of the input signal. This proportionality coefficient and the filter order were determined based on empirical validation, selecting the configuration exhibiting maximum reliability across all folds.
Importantly, the subsequent stages of the evaluation pipeline, including the peak-detection algorithm and the outlier-rejection controls, remained unmodified to ensure a fair and consistent comparison with all previous 32-bit floating-point models.}

\rev{It is worth noting that the choice of the models produced by each optimization step which are used as input for the subsequent ones, can be considered as an additional degree of freedom. We opted to use the most accurate points and the seeds, in order to primarily focus on obtaining accurate (yet compressed) DNNs; however, an alternative selection (e.g., taking multiple Pareto-optimal models from each curve) is also possible, with an obvious difference in the global optimization time.}

\subsection{Datasets}
We employ the same four datasets used in the most recent and comprehensive benchmarking study on PPG-based BP estimation \cite{nature-survey}. \rev{For details on each dataset and descriptive statistics, please refer to the original papers.}
\subsubsection{Sensors}
The Sensors \cite{sensors_dataset} dataset is a subset of the MIMIC-III database, containing records from 1,195 intensive care unit (ICU) patients. It includes demographic data along with PPG and ABP waveforms, collected using the Philips CareVue Clinical Information System and iMDsoft MetaVision ICU\cite{imdsoft-metavision-icu}. The dataset is the second largest in the set after UCI, with a total measurement duration of approximately 15 hours. Each record consists of two 15-second signal segments, spaced 5 minutes apart. 
\subsubsection{UCI}
The UCI \cite{uci_dataset} dataset, also known as the Cuff-Less Blood Pressure Estimation Dataset, is a subset of the MIMIC-II Waveform Database. Although MIMIC-II and MIMIC-III originate from the same underlying measurements, collected under identical conditions, in the same hospitals, and using the same devices, the UCI and SENSORS datasets are distinct and should not share records. However, UCI does not include subject information, making it impossible to check for data leakage across subjects.
Nevertheless, as the largest among the four, it is of particular interest for DL methods.
\subsubsection{BCG}
The BCG \cite{BCG_dataset} dataset is a bed-based ballistocardiography dataset collected from 40 patients, four of whom had preexisting heart conditions, while the rest were healthy. Data collection was conducted under Kansas State University IRB protocol \#9386, using the Finapres Medical Systems Finometer PRO for continuous brachial BP measurement and the GE Datex CardioCap 5 for PPG. The dataset is resampled from 1000 Hz to 125 Hz, and BP signals are rescaled by a factor of 100 mmHg/Volt. 
With a total of approximately 4 hours of recorded measurements, BCG is smaller than the UCI and SENSORS datasets. However, it has a notably high ratio of segments per subject, though its limited number of participants results in lower overall data variability.
\subsubsection{PPGBP}
PPGBP \cite{PPGB_dataset} dataset is the smallest among the four, totaling less than an hour of recordings. However, it includes 219 subjects, all with various cardiovascular diseases such as hypertension and diabetes. Blood pressure measurements were taken using the Omron HEM-7201 device, providing only SBP and DBP discrete values.
Following a 10-minute rest period, each subject underwent a single blood pressure measurement, followed by three 2.1-second PPG recordings using the SEP9AF-2 device. The original 1000 Hz signals were resampled to 125 Hz. With only three segments per subject, PPGBP has a notably low segment-to-subject ratio but exhibits high data variability relative to its small size.

\subsection{NAS Pareto Analysis}
 
\begin{figure}[ht]
    \centering
    \includegraphics[width=\textwidth]{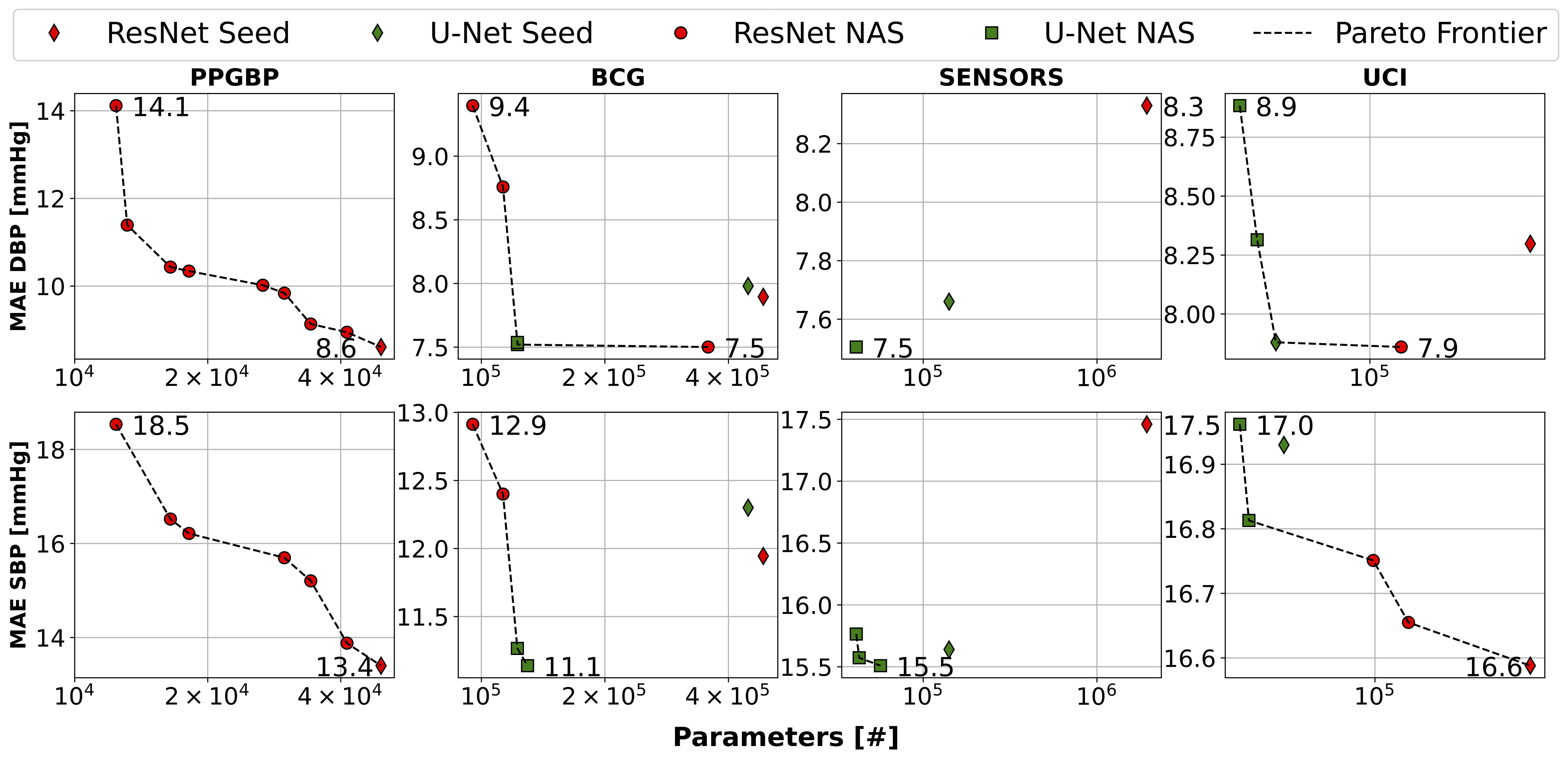}
    \caption{NAS results on all datasets on DBP (top row) and SBP prediction (bottom row).}
    \label{fig:res}
\end{figure}

Figure \ref{fig:res} depicts the results of NAS on all four datasets for both DBP and SBP prediction models. All results are reported in a MAE vs model size (n. of parameters) plane. Red and green diamonds correspond to seed DNNs (ResNet and U-Net from~\cite{nature-survey} respectively). Correspondingly colored circles and squares are the Pareto-optimal architectures found with NAS \rev{(the non-Pareto-optimal ones are omitted for visual clarity.)}

All results at this stage use floating point weights and activations.
While all our models simultaneously predict DBP and SBP, we report separate Pareto fronts for the two quantities for easier visualization.

On all datasets, we obtain models that either dominate the seeds or are on the memory vs error Pareto front. Namely, on PPGBP, we obtain a rich curve of Pareto architectures starting from the ResNet. 
We are able to reduce the seed size by 1.19$\times$, with a small increase in the MAE of only 3.9\% and 3.5\% on DBP and SBP prediction, respectively.
As mentioned earlier, U-Net-derived models are not applied to this dataset, since the full BP signal ground truth is not available.

On BCG, we Pareto-dominate both seed networks, improving both MAE and size.
Our most accurate U-Net-derived model obtains 11.139 mmHg MAE on SBP prediction and 7.52 mmHg MAE on DBP, being 6.7\%/4.7\% better than the best seed (ResNet). Simultaneously, this network reduces the total number of parameters by 3.8$\times$.

Similarly, on the two largest datasets, thanks to our NAS, we are again able to obtain Pareto-dominant solutions. On Sensors, our U-Net-derived architectures reduce the size of the most accurate seed (U-Net) by 2.5$\times$, while achieving a similar or lower MAE of 7.51 mmHg / 15.51 mmHg on DBP/SBP, respectively.  The situation reverses in UCI, where ResNet-derived DNNs achieve the best performance. The most accurate NAS output networks require only 149.8k/156.3k parameters to achieve a close-to-optimal MAE of 16.65 mmHg on SBP estimation, and the lowest overall MAE (7.86 mmHg) on DBP estimation. 
While the seed ResNet is able to achieve an even lower MAE on SBP, with its 792k parameters, it would be impossible to deploy on GAP8's internal memory of 512KB, even when quantized.

Interestingly, on BCG and Sensors, U-Net-based architectures outperform ResNets. We attribute this behavior to the ability of this network topology to learn faster from a lower amount of data, thanks to the richer training signal provided by the full-time series reconstruction task.

\rev{Lastly, we note that, already after the NAS stage, most of the obtained models have memory footprints compatible with the constraints of our target platform (512kB on-chip memory)~\cite{gap8}. As mentioned, this is the primary goal to ensure deployability. However, the following sections demonstrate how  the combined application of pruning and quantization further improves the Pareto front, yielding smaller models of similar accuracy, which would not only fit in even tighter memory constraints, but also execute faster and more efficiently, thanks to the reduced data movement and FLOPs, and to the integer data format better suited to the FPU-less GAP8.}

\subsection{Pruning Pareto analysis}\label{sub:pit}

\begin{figure*}[ht]
    \centering
    \includegraphics[width=\textwidth]{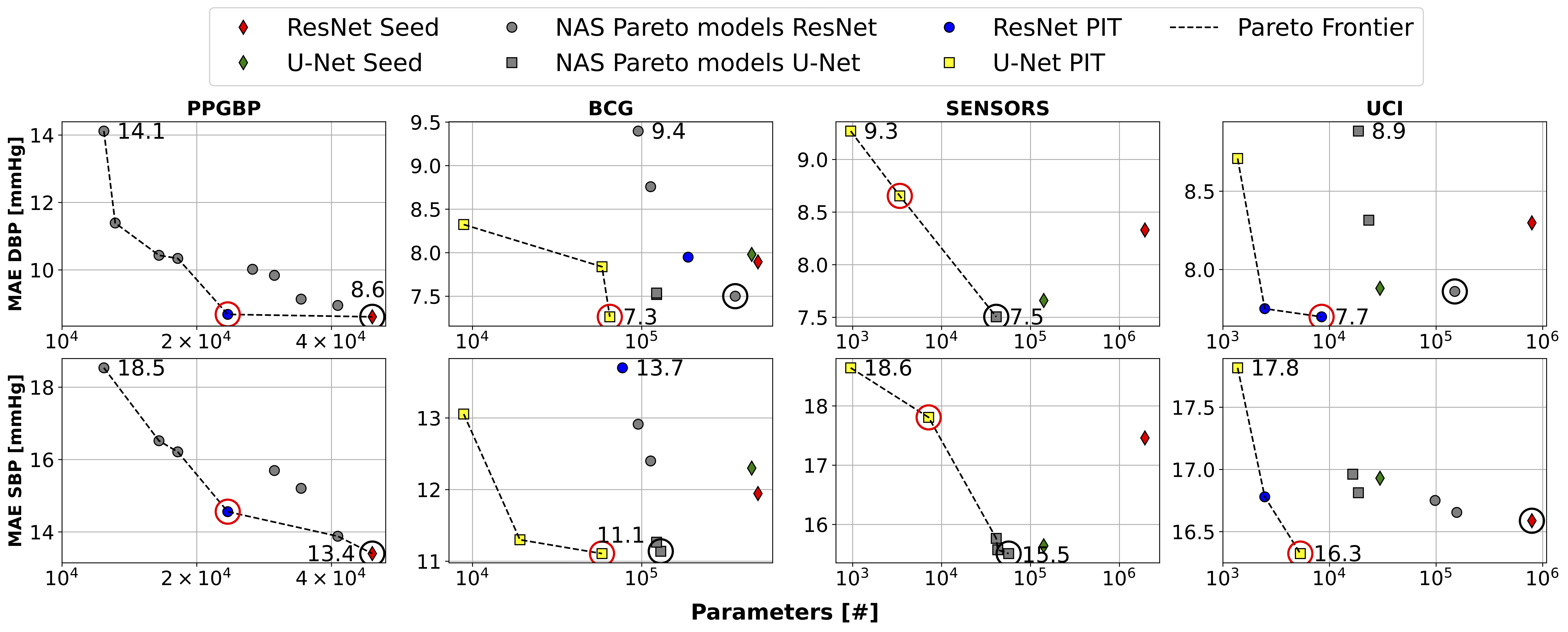}
    \caption{Pruning results on all datasets on DBP (top row) and SBP (bottom row) prediction. \rev{Black circles are PIT seeds, red circles are the models discussed in Section~\ref{sub:pit}.}}
    \label{fig:res_pit}
\end{figure*}

Figure \ref{fig:res_pit} shows that structured pruning further improves the Pareto front on all datasets, and generates new most-accurate models on two of them.  
\rev{
Starting from the previously obtained NAS Pareto front, we apply PIT to the \rev{black} circled models. Figure \ref{fig:res_pit} includes all points already shown in Figure~\ref{fig:res} (the seeds in red and green and the NAS Pareto-front in grey). Additionally, if Pareto-optimal, it also includes (in yellow and blue) the new results obtained by PIT. As before, intermediate (non-Pareto-optimal) pruning candidates are omitted for clarity.
}
\rev{The models circled in red are the ones that will be discussed in detail in this section.}

On PPGBP, PIT generates a new Pareto-optimal model starting from the seed ResNet.
Notably, the latter still achieves the lowest MAE (8.61mmHg vs 8.68mmHg), but at the cost of a 2.1$\times$ larger memory footprint with respect to the pruned model\rev{, outlined in red in the top-left chart of Fig. \ref{fig:res_pit}}.

On BCG, the new PIT models dominate all previous results, completely redefining the Pareto front. The most accurate models \rev{(red circles)} achieve a 7.99\% and 7.01\% decrease in DBP and SBP MAE, respectively, with a 7.5$\times$ and 8.4$\times$ reduction in the number of parameters compared to the ResNet seed. Alternatively, a 30.3$\times$ (13.3$\times$) decrease in model size is achieved while still improving the SBP (DBP) MAE by 1.31\% (1.77\%).

On Sensors, PIT created several models with drastically reduced memory footprint but also slightly larger errors. Namely, the most accurate PIT results, \rev{circled in red} achieved an 88.3$\times$ (70.9$\times$) parameters reduction with an increase in SBP (DBP) MAE of 1.73\% (1.47\%) compared to the ResNet seed and 14.53\% (12.6\%) with respect to the best NAS output.

On UCI, the biggest and most complex dataset, PIT achieved good accuracy, completely redefining the Pareto front and generating new most accurate models on both SBP and DBP. 
The most accurate PIT model \rev{marked in red in the lower rightmost square}, derived from U-Net, achieved a 1.59\% error reduction with 149$\times$ fewer weights on SBP compared to the previously most accurate ResNet seed. Similarly, on DBP, the most accurate PIT model (a ResNet) achieves a 2.3\% error reduction with 94$\times$ fewer parameters compared to the most accurate NAS output.

\subsection{Quantization and Mixed Precision Search Pareto analysis}\label{sub:mps}

\begin{figure}[ht]
    \centering
    \includegraphics[width=\textwidth]{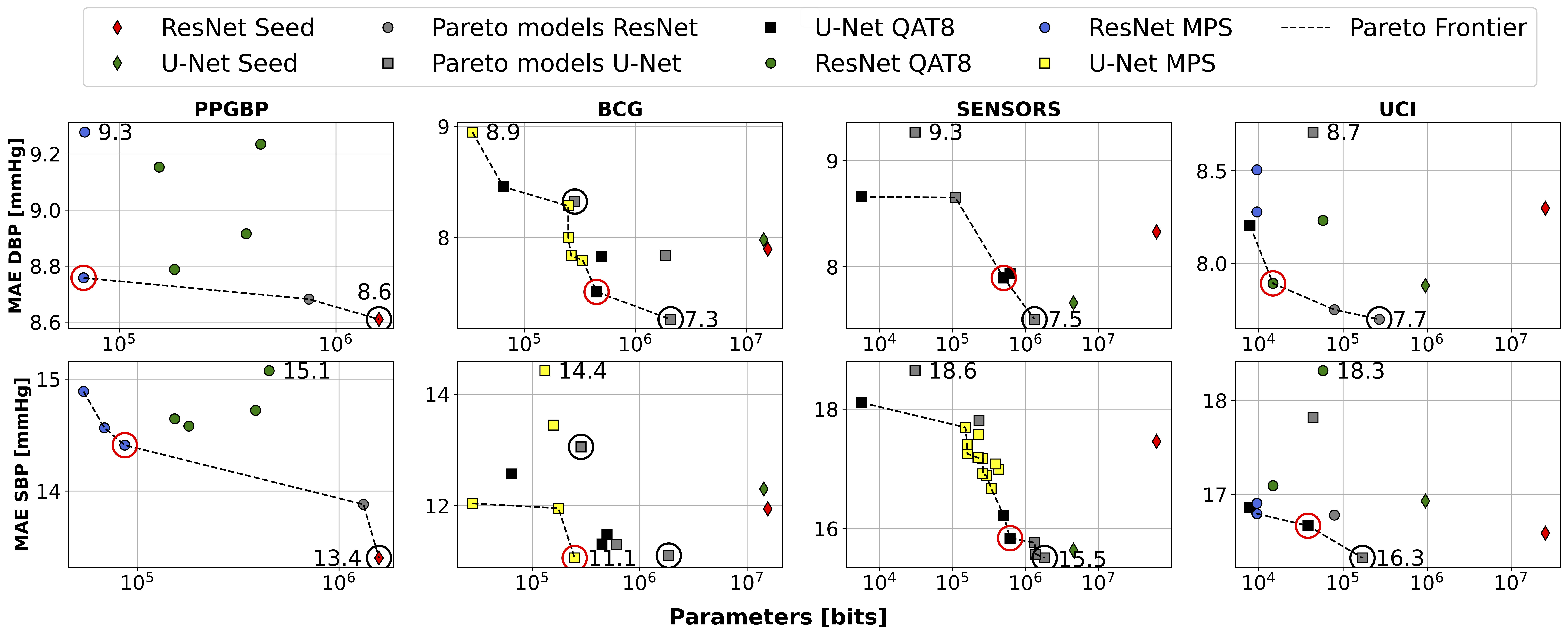}
    \caption{Results of MPS on all datasets on DBP (top row) and SBP (bottom row) prediction. \rev{Black circles are MPS seeds, red circles are the models discussed in Section~\ref{sub:mps}.}}
    \label{fig:res_mps}
\end{figure}

Figure~\ref{fig:res_mps} shows how the Pareto front further changes when applying the last step of our optimization chain, i.e., QAT/MPS. In this case, gray points represent the combined Pareto front of the NAS and pruning phases. As before, QAT/MPS have been applied to the models \rev{circled in black. Those circled in red, instead, are the ones discussed in the rest of this section in detail}. 
\rev{For clearer visualization of the new results, some of the least accurate models generated by the previous steps (NAS and pruning) have been omitted from the figure.}
The x-axis reports the total number of \textit{bits} required by each model instead of the number of parameters, to meaningfully compare networks using different data precision.
Quantization further advances the Pareto Front, creating models with slightly higher errors but a way smaller memory footprint.

On PPGBP, the most accurate quantized models show a 6.18\% and 8.25\% higher errors for DBP and SBP respectively, but requires 9.99$\times$ less memory bits, compared to the original ResNet.
MPS models improve significantly the left part of the Pareto front, \rev{yielding better accuracies compared to previous NAS+pruning outputs of similar size.}
On BCG our quantization algorithms were applied on U-Net models already optimized with both NAS and pruning. 
The most accurate DBP model, \rev{ red-circled in the second top square of Figure \ref{fig:res_mps},} obtains a 6.2$\times$ parameter reduction at the cost of a 6.95\% higher error compared to the most accurate previous network.
Most notably, on SBP, MPS yields a model with both 0.39\% lower error and a 7.5$\times$ smaller footprint than previous results. The lower error is probably due to the regularization effect of quantization on such a small architecture.

On SENSORS two U-Nets generated by NAS and one generated by PIT underwent quantization. The most accurate results were obtained by 8-bit models \rev{(the black squares circled in red in the SENSORS plots)}, e.g.,  5.2\% and 2.13\% higher DBP and SBP MAE, respectively, in exchange for a 4$\times$ memory reduction compared to the previous best model.

On the biggest dataset, UCI, we quantized both ResNet and U-Net based architectures, already optimized by the previous algorithms.
The most accurate quantized model for DBP prediction is an 8-bit precision ResNet, \rev{(red-circled green dot)}, which is 5.37$\times$ smaller and achieves a 2.52\% higher error, compared to the most accurate model overall. With regard to SBP, instead, the most accurate model is U-Net based, 8-bit quantized, and reaches a 2.1\% higher MAE than the best float solution, while requiring 4.44$\times$ fewer bits.

In conclusion, although the quantized models expectedly did not achieve substantially lower errors compared to floating point models, they provide significant memory reductions, and consequently energy savings once deployed, often with a negligible compromise in accuracy.

\subsection{Subject-specific Fine-tuning}
\label{sec:res_fine_tuning}

The key dataset requirements to assess the benefits of subject-specific fine-tuning are the availability of subject-level annotations 
and a sufficiently large number of samples per patient to enable both fine-tuning and evaluation. 
Based on this observation, we perform our fine-tuning experiments on BCG, since UCI does not provide subject identity metadata, and PPGBP and Sensors have only 3 and 9 samples per subject, respectively (versus approximately 75 in BCG).

\begin{figure}
    \centering
    \includegraphics[width=0.8\columnwidth]{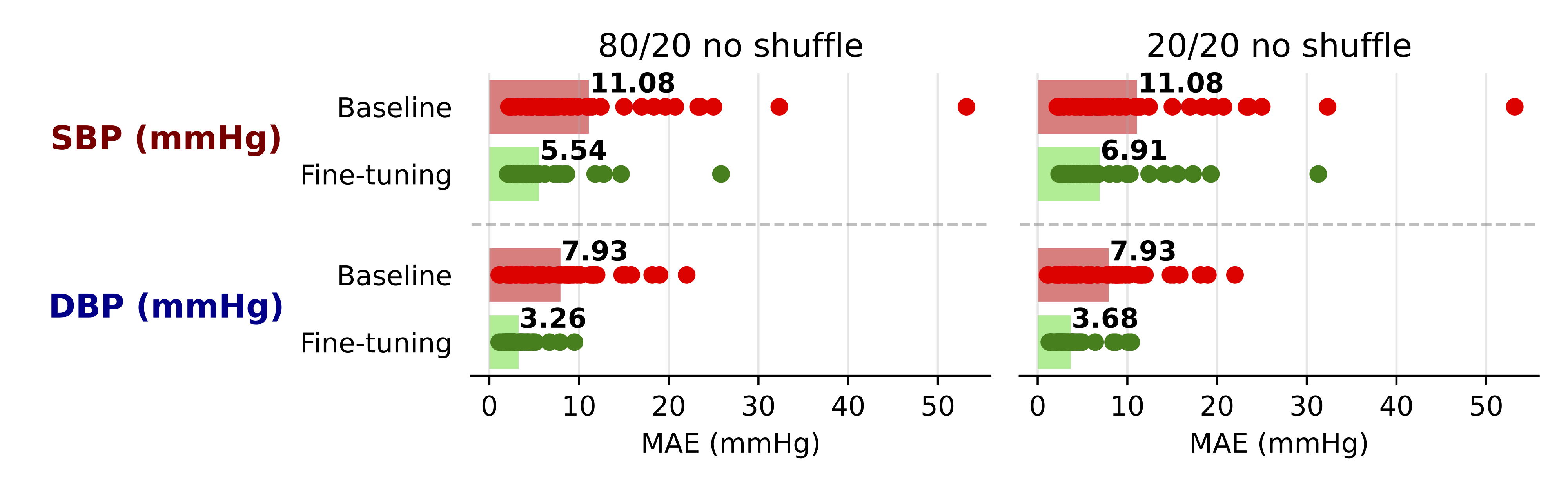}
    \caption{\rev{MAE comparison pre- and post- fine-tuning, without data shuffling, using 80\% of the samples as training set and 20\% test set (left), or 20\% as training set and 20\% as test set (right)}}\label{fig:fine-tuning-noshuffle}
\end{figure}

\begin{figure}
    \centering
    \includegraphics[width=0.8\columnwidth]{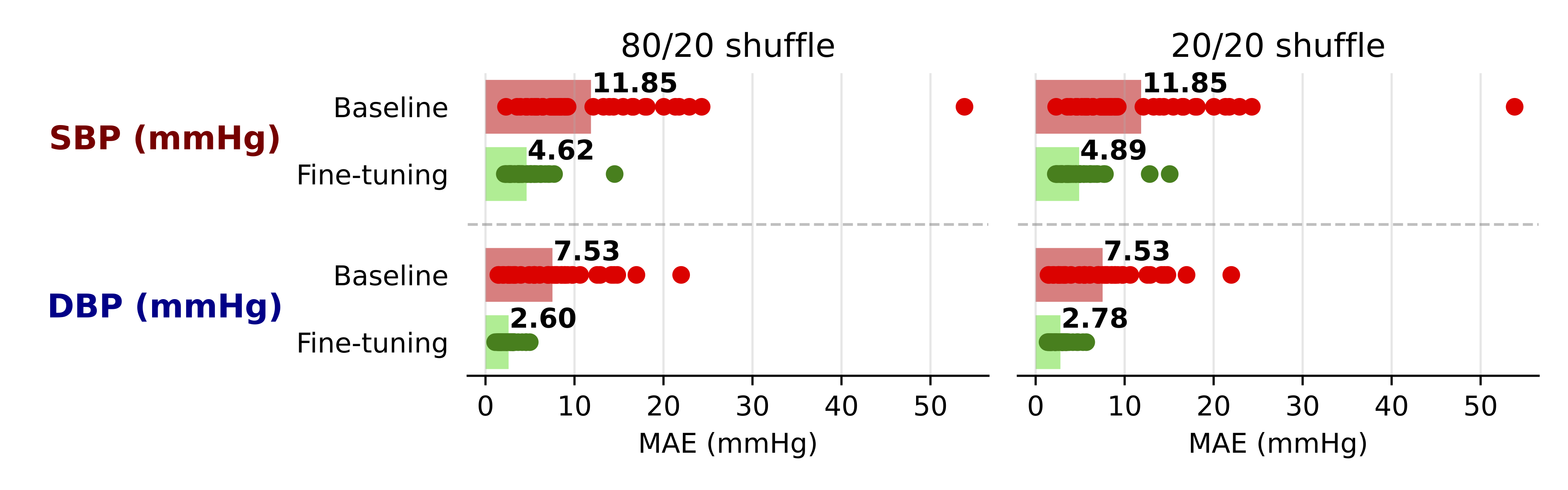}
    \caption{\rev{MAE comparison pre- and post- fine-tuning, with data shuffling, using 80\% of the samples as training set and 20\% test set (left), or 20\% as training set and 20\% as test set (right)}}
    \label{fig:fine-tuning-shuffle}
\end{figure}

We selected three quantized models optimized by our cross-subject pipeline (see previous sections), namely the best performing quantized models for SBP and DBP prediction, and the one with the least number of parameters. Fig.~\ref{fig:fine-tuning-noshuffle} and Fig. \ref{fig:fine-tuning-shuffle} show the MAE distribution before and after fine-tuning for one of these models (the one with the lowest DBP error). The other configurations show similar results, and their global MAE improvements after fine-tuning are reported in Table~\ref{tab:res-deployment}. Each dot in the graphs represents the MAE on one subject, while bars are the global averages. The difference between the two graphs is that Fig.~\ref{fig:fine-tuning-noshuffle} uses a \rev{\textit{contiguous chunk}} of the samples as a fine-tuning set (i.e., without shuffling $P_{test}$).
\rev{We keep samples in order} to maximize fairness, under the assumption that samples are temporally ordered in the dataset. \rev{Therefore, this split mechanism} corresponds to evaluating the fine-tuned model at a separate time with respect to its training. However, since the temporal ordering of samples is not clearly specified in the BCG metadata, we also experiment shuffling $P_{test}$ (Fig.~\ref{fig:fine-tuning-shuffle}).
\rev{In both figures, the leftmost graph refers to a setup that uses the entire set of samples for the subject under consideration, splitting them into 80\% for training and 20\% for testing (either temporally, in Fig.~\ref{fig:fine-tuning-noshuffle}, or randomly, in Fig.~\ref{fig:fine-tuning-shuffle}). Additionally, to assess the effectiveness of fine-tuning in a small-data setup, we repeat the experiment keeping the same test samples, but reducing the training set to a (disjoint) set of the same size, i.e., 20\% of the total, corresponding to 15 samples per subject on average. This experiment is useful to validate whether models could be effectively personalized through a reasonably short calibration session.}

In \rev{all 4} cases, fine-tuning demonstrates a pronounced positive effect on accuracy, providing a consistent reduction in MAE for both SBP and DBP. 
\rev{As expected, the lowest error is achieved when finetuning on the whole 80\% training set, both with and without shuffling.}
Compared to the baseline, fine-tuning without shuffling reduces the MAE by 50\% for SBP and by 58.9\% for DBP. More specifically, the MAE reduces for 36 out of the 40 subjects in both SBP and DBP. For the latter, the 36 improving subjects show an average MAE reduction of 5.51\,mmHg, whereas the remaining 4 experience an average increase of just 1.16\,mmHg. Similarly, for SBP, the average MAE improvement is 8.10\,mmHg, while the average degradation is just 0.62\,mmHg.
Shuffling data prior to fine-tuning results in
even greater improvements, with global MAE reductions of 61.01\% for SBP and 64.27\% for DBP. 

\rev{Additionally, the right side of the two figures shows that even
finetuning on a much smaller dataset yields consistent improvements. The SBP (DBP) MAE is reduced by 37.6\% (53.6\%) in the no-shuffling case, and by 58.7\% (63.1\%) in the shuffling case. This demonstrates that that competitive performance can be achieved even in tightly data-constrained fine-tuning setups.
}

\rev{Lastly, we note that no statistically significant difference was observed in the performance gains achieved through fine-tuning between the cohort of 36 healthy patients and the subset of 4 patients with cardiovascular diseases within the BCG dataset.}

These findings support the importance of subject-specific fine-tuning in improving BP prediction accuracy.

\subsection{Comparison with the state-of-the-art}\label{sec:sota}
\rev{
Table \ref{tab:sota} compares the results of our best performing models, without per-subject fine-tuning, against the best performing solutions (either classical models or DNNs) from the state-of-the-art, limited to those using the standardized preprocessing introduced in~\cite{nature-survey} for fairness of comparison.}

\begin{table}
\caption{Comparison with State of the Art methods from \cite{nature-survey} and \cite{Lim2025}.}
\centering
\label{tab:sota}
\begin{tabular}{|c|c c|c c|}
\hline
\multirow{3}{*}{Model} & \multicolumn{2}{c|}{SBP} & \multicolumn{2}{c|}{DBP} \\ \cline{2-5}
 & MAE & Params & MAE & Params \\ 
 & [mmHg] & [\#] & [mmHg] & [\#] \\ \hline

\noalign{\vskip 2pt}
\multicolumn{5}{c}{\textbf{PPGBP}} \\
\noalign{\vskip 2pt}
\hline
Random Forest              & 13.17 & 20.3k  & 8.12 & 20.4k  \\ \hline
Support Vector Regression  & \textbf{13.15} & 29.0k  & \textbf{8.04} & 16.9k  \\ \hline
Conv-Transformer \cite{Lim2025} & 14.82 & - & 9.17 & - \\ \hline 
Ours                       & 13.40 & 49.3k  & 8.61 & 49.3k  \\ \hline

\noalign{\vskip 2pt}
\multicolumn{5}{c}{\textbf{BCG}} \\
\noalign{\vskip 2pt}
\hline
Random Forest              & 12.88 & 0.21k   & 7.89 & 85.3k  \\ \hline
Support Vector Regression  & 11.45 & 10.7k   & 7.34 & 55.9k  \\ \hline
V-Net & 11.42 & 491k & 8.01 & 491k \\ \hline
Ours                       & \textbf{11.07} & 58.2k  & \textbf{7.26} & 64.8k  \\ \hline

\noalign{\vskip 2pt}
\multicolumn{5}{c}{\textbf{Sensors}} \\
\noalign{\vskip 2pt}
\hline
Random Forest              & 15.86 & 64.0k   & 7.66 & 171k   \\ \hline
Support Vector Regression  & 15.60 & 775k    & \textbf{7.50} & 416k   \\ \hline
PPGBIABP & 16.45 & 92.5M & 7.99 & 92.5M\\ \hline
Ours                       & \textbf{15.51} & 56.8k  & \textbf{7.50} & 41.2k  \\ \hline

\noalign{\vskip 2pt}
\multicolumn{5}{c}{\textbf{UCI}} \\
\noalign{\vskip 2pt}
\hline
Random Forest              & 16.85 & 21.3k   & 8.25 & 4.26k  \\ \hline
Support Vector Regression  & 17.45 & 18.5M   & 8.07 & 4.54M  \\ \hline
PPGBIABP & 17.06 & 296k & 8.07 & 296k\\ \hline
Ours                       & \textbf{16.32} & 5.32k  & \textbf{7.69} & 8.43k  \\ \hline

\end{tabular}
\end{table}

\rev{Results show that, on 3 out of 4 datasets, our automated DNN optimization pipeline yields a model with lower error than the SotA. The only exception is PPGBP, where classical ML models (SVR and RF) outperform our optimized DNNs in both performance and size. 
Namely, the most accurate of the two, i.e., the SVR,
achieves a MAE of 8.04mmHg and 13.15mmHg in DBP and SBP estimation, respectively. In comparison, our most accurate DNNs obtain 8.61 and 13.40, i.e., a 7\% and 2\% increase. 
The SVRs are also more parameter-efficient, being 2.42x (2.93x) smaller than our most accurate DNNs for SBP (DBP).}
\rev{We impute this result to the small size of this dataset, which favors classical models over DNNs.}

\rev{On BCG, classical models remain competitive, although our DNNs obtain slightly lower error, namely 3.7\% (1.1\%) for SBP (DBP), at the cost of a significant size increase for SBP (5.44x) and a moderate one for DBP (1.16x). However, it shall be remarked that~\cite{nature-survey} trains \textit{separate} SVR and RF models for SBP and DBP prediction. Therefore, when considering a full ABP monitoring system that predicts both values, those approaches would require two separate models, which makes our DNNs both more accurate and more compact, as detailed in Section~\ref{sec:deployment}. Additionally, compared to the best performing DNNs for SBP (V-Net), our model achieves a slight accuracy improvement, with a huge parameter reduction of 7.5$\times$.}

\rev{On Sensors, classical ML methods, that were shown to outperform the original U-Net and ResNet seeds in~\cite{nature-survey}, are instead surpassed by our NAS-optimized DNNs. Namely, SVRs, which achieved the best results on both metrics, are now matched by our U-Net NAS models in terms of error (0.6\% lower for SBP and similar for DBP), while our models also require 13.7$\times$ (10.1$\times$) fewer parameters. Comparing our best model with one of the best DNNs (excluding ResNet and U-Net, which were previously considered), PPGBIAP, we are able to improve the performance in both metrics with an impressive 2250$\times$ parameter reduction.}

\rev{Lastly, on UCI, the dataset with the largest number of samples, our models strongly outperform both the SotA DNNs and classic methods, achieving 6.5\% (4.7\%) lower MAE on SBP (DBP) and reducing the required number of parameters by a striking 3500$\times$ (540$\times$). In fact, the higher complexity of this dataset causes the number of parameters of both the SVR (with RBF kernel) and the RF to increase exponentially, further advocating for a DNN-based approach.}

\subsection{Network Deployment}\label{sec:deployment}

Figure~\ref{fig:res_mps} shows that Pareto-optimal (quantized) models for all datasets have a limited memory footprint, lower than the 512kB available on-board our target platform (GAP8). Notably, this means that  all of them could theoretically be deployed.

However, in this section, we focus on deployment results for models trained on BCG, i.e, the same dataset on which we experimented with subject-specific fine-tuning (see Sec.~\ref{sec:res_fine_tuning}). 
\rev{Namely, we deploy three neural networks optimized with our proposed pipeline, as well as three models from the state-of-the-art~\cite{nature-survey}. We select the same (best) baselines reported in Table~\ref{tab:sota} for BCG, which include two classic ML algorithms, a RF and a SVR, and a signal-to-signal VNet DNN. For what concerns our results, we select the two models with the lowest SBP and DBP MAE (among quantized ones) and in addition to that, the smallest Pareto-optimal model overall.}

Table \ref{tab:res-deployment} reports the inference costs of these \rev{models}, in particular: the memory footprint, the total number of MAC operations, and the latency and energy consumption per input window. Results refer to GAP8 operating at 100MHz. Latency values are read from performance counters, while energy is estimated considering an average active power consumption of 51mW. We also report the DBP and SBP MAEs before and after subject-specific fine-tuning.
It's worth noting how the best SBP model before finetuning has a slightly higher SBP MAE than the best DBP model (4.70 against 4.62). That could be explained by the smaller size of the former, which makes it less capable of improvement with subject-specific fine-tuning. The "Best SBP" model is also a mixed precision one, while "Best DBP" uses all 8-bit weights, which explains why it is faster to execute despite a larger memory occupation.

\begin{table*}[ht]
\centering
\caption{Deployment results on GAP8.}

\label{tab:res-deployment}
\footnotesize
\begin{tabular}{lcccccccc}
\hline
Network & MACs [M] & Cycles & MACs/Cycles & DBP MAE & SBP MAE & Latency [ms] & Energy [mJ] & Memory [KByte]  \\\hline

\rev{RF (DBP + SBP)} & \rev{n.a.} & \rev{366.6K + 1.2M} & \rev{n.a.} & \rev{7.89} & \rev{12.88} & \rev{3.7 + 12.2} & \rev{0.19 + 0.62} & \rev{85.3 + 0.21}  \\

\rev{SVR (DBP + SBP)} & \rev{n.a.} & \rev{764.6K + 1.4M} & \rev{n.a.} & \rev{\textbf{7.34}} & \rev{11.45} & \rev{7.64 + 13.89} & \rev{0.39 + 0.71} & \rev{55.9 + 10.7}  \\

\rev{VNet}    & \rev{64.4} & \rev{6.7M}   & \rev{9.55} & \rev{8.01} & \rev{11.42} & \rev{67.49} & \rev{3.44} & \rev{491}  \\\hline
Best DBP      & 47.09      & 13.2M & 3.56 & 7.51 (2.68)$^{\dagger}$ & 11.31 (4.62)$^{\dagger}$ & 132.24 & 6.74 & 54.35  \\
Best SBP     & 48.59 & 24.1M & 2.01 & 8.28 (2.80)$^{\dagger}$ & \textbf{11.06} (4.70)$^{\dagger}$ & 241.33 & 12.31& 30.26     \\
Best Params & 11.61 & 5.3M  & 2.20 & 9.32 (7.72)$^{\dagger}$ & 12.25 (11.06)$^{\dagger}$ & 52.87 & 2.70 & 2.15  \\\hline
\end{tabular}
\begin{flushleft}
\footnotesize $^{\dagger}$ Values in parentheses come from fine-tuning and are computed on a slightly different set of data. For further details on the fine-tuning protocol please refer to Section \ref{sec:fine_tuning}
\end{flushleft}
\end{table*}

\rev{As anticipated in Section~\ref{sec:sota} classical models (SVR and RF) are \textit{separately trained} for SBP and DBP prediction. Thus, a complete ABP monitoring system based on these approaches would require the deployment of two separate models. Accordingly, we report in Table~\ref{tab:res-deployment} the memory, energy and latency results for both sub-models, which must be summed to obtain the totals (assuming sequential execution). Conversely, the VNet, like our models, can be used to predict both SBP and DBP.}

\rev{The results show that our optimized models outperform all baselines in terms of memory occupation versus accuracy trade-off. Even the largest one occupies merely 54.4 kB, and the smallest one remarkably only requires 2.15 kB of memory,} while still achieving decent MAE results of 9.32 (7.72) for DBP and 12.25 (11.06) for SBP before and after fine-tuning. \rev{In comparison, the RF and SVR require a total of 66.6 kB and 85.5kB respectively. The VNet requires even more space (491kB), exceeding the GAP8 L2's 512kB (which also contains code segments, temporary data, etc), and requiring the use of external memory, connected via SPI, for storing weights.}

\rev{In terms of latency and energy consumption, classic ML models are superior to DNNs, as expected, given the nature and number of the involved operations per prediction (e.g., the RF is approximately 3.35$\times$ faster and more efficient than our smallest DNN). Our results are instead comparable to those of VNet, which achieves slightly lower latency with respect to our most accurate models thanks to the usage of normal convolutions, as opposed to harder to accelerate (but more parameter-efficient) depthwise layers.}
\rev{Importantly, however, all our models largely meet real-time performance constraints,} as the inference latency ranges between 52.87 and 241.33 ms, thus being significantly lower than the time duration of a BCG input window, i.e., 5 seconds.

\rev{While not more efficient than all baselines, our networks still allow days of continuous monitoring on a wearable without recharging. In fact, considering their consumption, which ranges between 12.31 and 2.7 mJ per inference, and} assuming a typical wearable battery with 500mAh capacity @3.7V, \rev{our models} would allow roughly between 500k and 2.5M inferences on a single charge, translating to 30 and 150 days of continuous monitoring. These simple calculations only consider processing energy, which is clearly an approximation. However, note that our DNN's energy consumption results are lower than those reported in~\cite{q-ppg} where a full-system analysis is performed for a similar PPG-based application.

\begin{figure}[ht]
    \centering
    \includegraphics[width=0.9\linewidth]{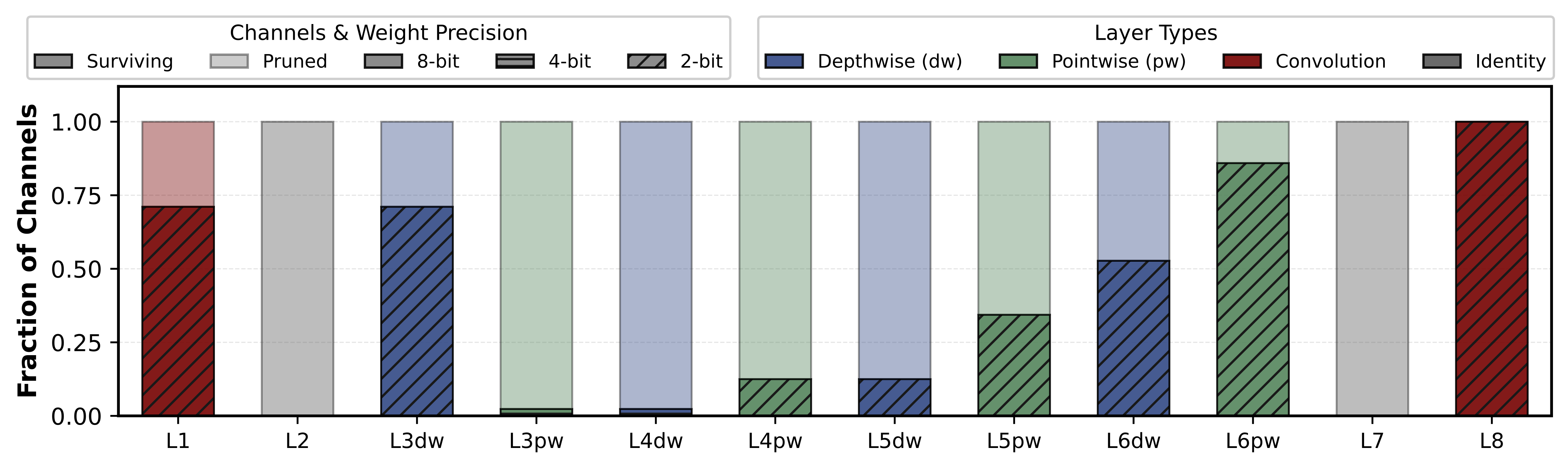}
    \caption{Visualization of the choices made by the pipeline during the optimization of the 'Best Params' deployed model} 
    \label{fig:opt_hist_params}
\end{figure}

\begin{figure}[ht]
    \centering
    \includegraphics[width=0.9\linewidth]{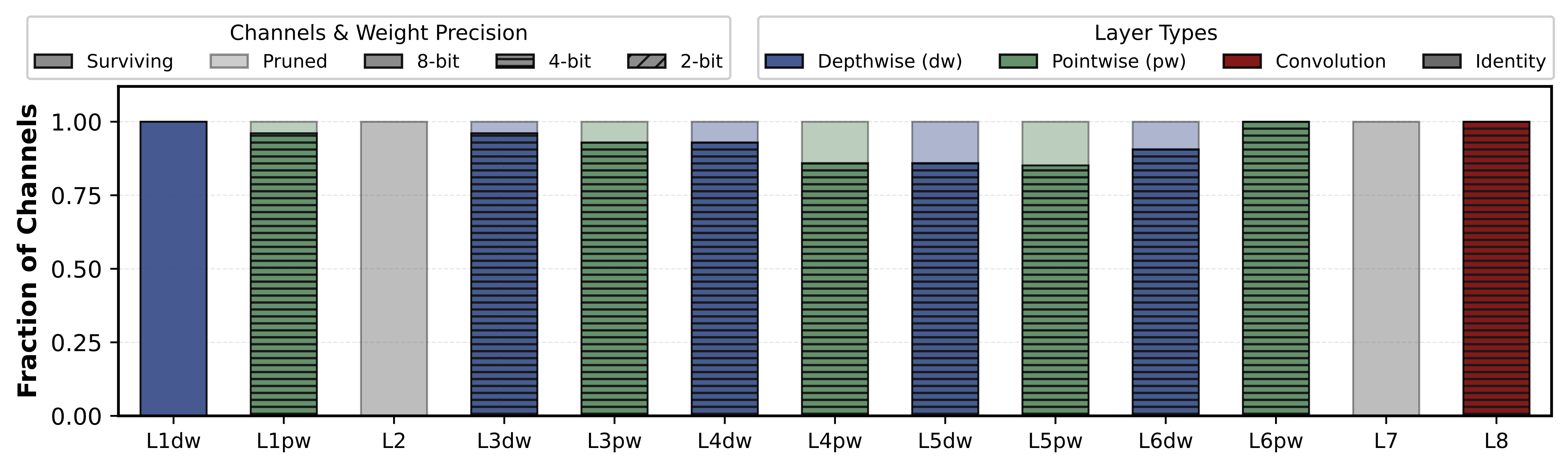}
    \caption{Visualization of the choices made by the pipeline during the optimization of the 'Best SBP' deployed model} 
    \label{fig:opt_hist_sbp}
\end{figure}

\subsection{Architectural Analysis}\label{sec:archi_analysis}
\rev{Figures \ref{fig:opt_hist_params} and \ref{fig:opt_hist_sbp} provide a layerwise view of two of the three deployed models, \emph{Best Params} and \emph{Best SBP} from Table \ref{tab:res-deployment}. We omit \emph{Best DBP} since it is a fully-8-bit DNN, thus not allowing an analysis of the MPS results. In each plot, every bar corresponds to one network layer: the \textbf{color} encodes the operator selected by SuperNet, the \textbf{height} indicates the fraction of channels retained after PIT pruning, and the \textbf{hatching} denotes the weight bit-width assigned by MPS.}
\rev{Across both deployed networks, SuperNet shows a strong preference for \emph{depthwise-separable convolutions}, suggesting that this operator is particularly well matched to PPG processing within our search-space constraints, while not being present in the baseline architectures. Moreover, several layers near the input and output are replaced by \emph{identity} mappings (i.e., fully bypassed), indicating that a shallower encoder/decoder is sufficient to capture and reconstruct the relevant information from PPG signals, and that additional depth in these regions would be largely redundant.}
\rev{The pruning behavior learned by PIT differs markedly between the two deployments, consistently reflecting the cost term controlled by $\lambda$. In \emph{Best Params} (Fig.~\ref{fig:opt_hist_params}), optimized with a larger $\lambda$ to emphasize efficiency, PIT performs substantial channel reduction across most layers. In contrast, \emph{Best SBP} (Fig.~\ref{fig:opt_hist_sbp}), optimized to prioritize estimation accuracy, retains a considerably larger portion of channels overall. Beyond this global trade-off, the pruning decisions reveal a consistent \emph{layerwise} pattern in both models: intermediate layers are pruned more aggressively, whereas early and late layers preserve higher channel counts. This suggests that (i) early layers are more critical for robust feature extraction from raw PPG and (ii) later layers are more sensitive due to their role in reconstruction/prediction, while mid-network representations contain more removable redundancy.}
\rev{Finally, MPS assigns mixed-precision in a largely even manner across layers, while still differentiating the two operating points. The \emph{Best Params} model is dominated by \emph{2-bit} weight assignments, enabling more extreme compression under the efficiency-driven objective. Conversely, the \emph{Best SBP} model uses \emph{4-bit} precision in several layers, preserving accuracy while still providing significant quantization benefits.}

\section{Conclusion}
Continuous blood pressure (BP) monitoring is crucial for the early detection and prevention of cardiovascular diseases. While most research in this field has focused on maximizing monitoring accuracy, an equally important but often overlooked aspect is the ability to deploy these models on ultra-low-power SoCs suitable for wearable applications, thus enabling low-cost continuous monitoring. In this work, we presented a fully automated pipeline for the optimization of DNNs for PPG-based BP prediction consisting of chained NAS, pruning, and mixed-precision-search steps. Each stage contributes to improving either the accuracy or the efficiency of the models, thus moving the Pareto front between performance and resource usage. From this process, we selected a subset of optimized models for deployment on the GAP8 ultra-low-power SoC, and evaluated them in terms of inference latency, energy consumption, and memory footprint, showing that our models enable multiple weeks of real-time continuous monitoring on battery-powered wearables. Additionally, we investigated patient-specific fine-tuning, an approach increasingly adopted by companies developing wearable devices, to further enhance model accuracy for individual users. \rev{Future work will explore adaptive on-device learning and multi-sensor data fusion to further improve long-term personalization and robustness in dynamic real-world conditions, for example by tackling motion artifacts.}

\bibliographystyle{ACM-Reference-Format}
\bibliography{acmart}


\begin{thebibliography}{70}


\ifx \showCODEN    \undefined \def \showCODEN     #1{\unskip}     \fi
\ifx \showISBNx    \undefined \def \showISBNx     #1{\unskip}     \fi
\ifx \showISBNxiii \undefined \def \showISBNxiii  #1{\unskip}     \fi
\ifx \showISSN     \undefined \def \showISSN      #1{\unskip}     \fi
\ifx \showLCCN     \undefined \def \showLCCN      #1{\unskip}     \fi
\ifx \shownote     \undefined \def \shownote      #1{#1}          \fi
\ifx \showarticletitle \undefined \def \showarticletitle #1{#1}   \fi
\ifx \showURL      \undefined \def \showURL       {\relax}        \fi
\providecommand\bibfield[2]{#2}
\providecommand\bibinfo[2]{#2}
\providecommand\natexlab[1]{#1}
\providecommand\showeprint[2][]{arXiv:#2}

\bibitem[688(2014)]%
        {6882122}
 \bibinfo{year}{2014}\natexlab{}.
\newblock \showarticletitle{IEEE Standard for Wearable Cuffless Blood Pressure Measuring Devices}.
\newblock \bibinfo{journal}{\emph{IEEE Std 1708-2014}} (\bibinfo{year}{2014}), \bibinfo{pages}{1--38}.
\newblock
\href{https://doi.org/10.1109/IEEESTD.2014.6882122}{doi:\nolinkurl{10.1109/IEEESTD.2014.6882122}}


\bibitem[Aguirre et~al\mbox{.}(2021)]%
        {sensors_dataset}
\bibfield{author}{\bibinfo{person}{Nicolas Aguirre}, \bibinfo{person}{Edith Grall-Maës}, \bibinfo{person}{Leandro~J. Cymberknop}, {and} \bibinfo{person}{Ricardo~L. Armentano}.} \bibinfo{year}{2021}\natexlab{}.
\newblock \showarticletitle{Blood Pressure Morphology Assessment from Photoplethysmogram and Demographic Information Using Deep Learning with Attention Mechanism}.
\newblock \bibinfo{journal}{\emph{Sensors}} \bibinfo{volume}{21}, \bibinfo{number}{6} (\bibinfo{year}{2021}).
\newblock
\showISSN{1424-8220}
\href{https://doi.org/10.3390/s21062167}{doi:\nolinkurl{10.3390/s21062167}}


\bibitem[Alexandre et~al\mbox{.}(2023)]%
        {Alexandre2023}
\bibfield{author}{\bibinfo{person}{Jérémy Alexandre}, \bibinfo{person}{Kevin Tan}, \bibinfo{person}{Tiago~P. Almeida}, \bibinfo{person}{Josep Sola}, \bibinfo{person}{Bruce~S. Alpert}, {and} \bibinfo{person}{Jay Shah}.} \bibinfo{year}{2023}\natexlab{}.
\newblock \showarticletitle{Validation of the Aktiia blood pressure cuff for clinical use according to the ANSI/AAMI/ISO 81060-2:2013 protocol}.
\newblock \bibinfo{journal}{\emph{Blood Pressure Monitoring}}  \bibinfo{volume}{28} (\bibinfo{date}{4} \bibinfo{year}{2023}), \bibinfo{pages}{109}.
\newblock
Issue 2.
\showISSN{14735725}
\href{https://doi.org/10.1097/MBP.0000000000000639}{doi:\nolinkurl{10.1097/MBP.0000000000000639}}


\bibitem[Ali and Atef(2023)]%
        {ppg-lstm}
\bibfield{author}{\bibinfo{person}{Noor~Faris Ali} {and} \bibinfo{person}{Mohamed Atef}.} \bibinfo{year}{2023}\natexlab{}.
\newblock \showarticletitle{An efficient hybrid LSTM-ANN joint classification-regression model for PPG based blood pressure monitoring}.
\newblock \bibinfo{journal}{\emph{Biomedical Signal Processing and Control}}  \bibinfo{volume}{84} (\bibinfo{date}{7} \bibinfo{year}{2023}), \bibinfo{pages}{104782}.
\newblock
\showISSN{1746-8094}
\href{https://doi.org/10.1016/J.BSPC.2023.104782}{doi:\nolinkurl{10.1016/J.BSPC.2023.104782}}


\bibitem[Almeida et~al\mbox{.}(2023)]%
        {almeidaAktiia2023}
\bibfield{author}{\bibinfo{person}{Tiago~P. Almeida}, \bibinfo{person}{Meritxell Cort{\'e}s}, \bibinfo{person}{David Perruchoud}, \bibinfo{person}{J{\'e}r{\'e}my Alexandre}, \bibinfo{person}{Pascale Vermare}, \bibinfo{person}{Josep Sola}, \bibinfo{person}{Jay Shah}, \bibinfo{person}{Luisa Marques}, {and} \bibinfo{person}{Cyril Pellaton}.} \bibinfo{year}{2023}\natexlab{}.
\newblock \showarticletitle{Aktiia Cuffless Blood Pressure Monitor Yields Equivalent Daytime Blood Pressure Measurements Compared to a 24-h Ambulatory Blood Pressure Monitor: {{Preliminary}} Results from a Prospective Single-Center Study}.
\newblock \bibinfo{journal}{\emph{Hypertension Research}} \bibinfo{volume}{46}, \bibinfo{number}{6} (\bibinfo{date}{June} \bibinfo{year}{2023}), \bibinfo{pages}{1456--1461}.
\newblock
\showISSN{1348-4214}
\href{https://doi.org/10.1038/s41440-023-01258-2}{doi:\nolinkurl{10.1038/s41440-023-01258-2}}


\bibitem[Arjomand et~al\mbox{.}(2025)]%
        {arjomand2025transforhythmtransformerarchitectureconductive}
\bibfield{author}{\bibinfo{person}{Amir Arjomand}, \bibinfo{person}{Amin Boudesh}, \bibinfo{person}{Farnoush Bayatmakou}, \bibinfo{person}{Kenneth~B. Kent}, {and} \bibinfo{person}{Arash Mohammadi}.} \bibinfo{year}{2025}\natexlab{}.
\newblock \bibinfo{title}{TransfoRhythm: A Transformer Architecture Conductive to Blood Pressure Estimation via Solo PPG Signal Capturing}.
\newblock
\showeprint[arxiv]{2404.15352}~[eess.SP]
\urldef\tempurl%
\url{https://arxiv.org/abs/2404.15352}
\showURL{%
\tempurl}


\bibitem[Beh et~al\mbox{.}(2023)]%
        {10.1145/3587256}
\bibfield{author}{\bibinfo{person}{Win-Ken Beh}, \bibinfo{person}{Yu-Chia Yang}, \bibinfo{person}{Yi-Cheng Lo}, \bibinfo{person}{Yun-Chieh Lee}, {and} \bibinfo{person}{An-Yeu(Andy) Wu}.} \bibinfo{year}{2023}\natexlab{}.
\newblock \showarticletitle{Machine-aided PPG Signal Quality Assessment (SQA) for Multi-mode Physiological Signal Monitoring}.
\newblock \bibinfo{journal}{\emph{ACM Trans. Comput. Healthcare}} \bibinfo{volume}{4}, \bibinfo{number}{2}, Article \bibinfo{articleno}{14} (\bibinfo{date}{April} \bibinfo{year}{2023}), \bibinfo{numpages}{20}~pages.
\newblock
\href{https://doi.org/10.1145/3587256}{doi:\nolinkurl{10.1145/3587256}}


\bibitem[Benfenati et~al\mbox{.}(2024)]%
        {benfenati2024enhanceppgimprovingppgbasedheart}
\bibfield{author}{\bibinfo{person}{Luca Benfenati}, \bibinfo{person}{Sofia Belloni}, \bibinfo{person}{Alessio Burrello}, \bibinfo{person}{Panagiotis Kasnesis}, \bibinfo{person}{Xiaying Wang}, \bibinfo{person}{Luca Benini}, \bibinfo{person}{Massimo Poncino}, \bibinfo{person}{Enrico Macii}, {and} \bibinfo{person}{Daniele~Jahier Pagliari}.} \bibinfo{year}{2024}\natexlab{}.
\newblock \bibinfo{title}{EnhancePPG: Improving PPG-based Heart Rate Estimation with Self-Supervision and Augmentation}.
\newblock
\showeprint[arxiv]{2412.17860}~[eess.SP]
\urldef\tempurl%
\url{https://arxiv.org/abs/2412.17860}
\showURL{%
\tempurl}


\bibitem[Bian et~al\mbox{.}(2020)]%
        {9176231}
\bibfield{author}{\bibinfo{person}{Dayi Bian}, \bibinfo{person}{Pooja Mehta}, {and} \bibinfo{person}{Nandakumar Selvaraj}.} \bibinfo{year}{2020}\natexlab{}.
\newblock \showarticletitle{Respiratory Rate Estimation using PPG: A Deep Learning Approach}. In \bibinfo{booktitle}{\emph{2020 42nd Annual International Conference of the IEEE Engineering in Medicine \& Biology Society (EMBC)}}. \bibinfo{pages}{5948--5952}.
\newblock
\href{https://doi.org/10.1109/EMBC44109.2020.9176231}{doi:\nolinkurl{10.1109/EMBC44109.2020.9176231}}


\bibitem[Bruschi et~al\mbox{.}(2020)]%
        {pulp-nn-mixed}
\bibfield{author}{\bibinfo{person}{Nazareno Bruschi}, \bibinfo{person}{Angelo Garofalo}, \bibinfo{person}{Francesco Conti}, \bibinfo{person}{Giuseppe Tagliavini}, {and} \bibinfo{person}{Davide Rossi}.} \bibinfo{year}{2020}\natexlab{}.
\newblock \showarticletitle{Enabling mixed-precision quantized neural networks in extreme-edge devices}. In \bibinfo{booktitle}{\emph{Proceedings of the 17th ACM International Conference on Computing Frontiers}} (Catania, Sicily, Italy) \emph{(\bibinfo{series}{CF '20})}. \bibinfo{publisher}{Association for Computing Machinery}, \bibinfo{address}{New York, NY, USA}, \bibinfo{pages}{217–220}.
\newblock
\showISBNx{9781450379564}
\href{https://doi.org/10.1145/3387902.3394038}{doi:\nolinkurl{10.1145/3387902.3394038}}


\bibitem[Burrello et~al\mbox{.}(2024)]%
        {10798404}
\bibfield{author}{\bibinfo{person}{Alessio Burrello}, \bibinfo{person}{Francesco Carlucci}, \bibinfo{person}{Giovanni Pollo}, \bibinfo{person}{Xiaying Wang}, \bibinfo{person}{Massimo Poncino}, \bibinfo{person}{Enrico Macii}, \bibinfo{person}{Luca Benini}, {and} \bibinfo{person}{Daniele~Jahier Pagliari}.} \bibinfo{year}{2024}\natexlab{}.
\newblock \showarticletitle{Optimization and Deployment of Deep Neural Networks for PPG-based Blood Pressure Estimation Targeting Low-power Wearables}. In \bibinfo{booktitle}{\emph{2024 IEEE Biomedical Circuits and Systems Conference (BioCAS)}}. \bibinfo{pages}{1--5}.
\newblock
\href{https://doi.org/10.1109/BioCAS61083.2024.10798404}{doi:\nolinkurl{10.1109/BioCAS61083.2024.10798404}}


\bibitem[Burrello et~al\mbox{.}(2020)]%
        {dory}
\bibfield{author}{\bibinfo{person}{Alessio Burrello}, \bibinfo{person}{Angelo Garofalo}, \bibinfo{person}{Nazareno Bruschi}, \bibinfo{person}{Giuseppe Tagliavini}, \bibinfo{person}{Davide Rossi}, {and} \bibinfo{person}{Francesco Conti}.} \bibinfo{year}{2020}\natexlab{}.
\newblock \showarticletitle{{DORY:} Automatic End-to-End Deployment of Real-World DNNs on Low-Cost IoT MCUs}.
\newblock \bibinfo{journal}{\emph{CoRR}}  \bibinfo{volume}{abs/2008.07127} (\bibinfo{year}{2020}).
\newblock
\showeprint[arXiv]{2008.07127}
\urldef\tempurl%
\url{https://arxiv.org/abs/2008.07127}
\showURL{%
\tempurl}


\bibitem[Burrello et~al\mbox{.}(2022a)]%
        {burrello2022bioformers}
\bibfield{author}{\bibinfo{person}{Alessio Burrello}, \bibinfo{person}{Francesco~Bianco Morghet}, \bibinfo{person}{Moritz Scherer}, \bibinfo{person}{Simone Benatti}, \bibinfo{person}{Luca Benini}, \bibinfo{person}{Enrico Macii}, \bibinfo{person}{Massimo Poncino}, {and} \bibinfo{person}{Daniele~Jahier Pagliari}.} \bibinfo{year}{2022}\natexlab{a}.
\newblock \bibinfo{title}{Bioformers: Embedding Transformers for Ultra-Low Power sEMG-based Gesture Recognition}.
\newblock
\showeprint[arxiv]{2203.12932}~[eess.SP]


\bibitem[Burrello et~al\mbox{.}(2022b)]%
        {ppg-hr}
\bibfield{author}{\bibinfo{person}{Alessio Burrello}, \bibinfo{person}{Daniele~Jahier Pagliari}, \bibinfo{person}{Pierangelo~Maria Rapa}, \bibinfo{person}{Matilde Semilia}, \bibinfo{person}{Matteo Risso}, \bibinfo{person}{Tommaso Polonelli}, \bibinfo{person}{Massimo Poncino}, \bibinfo{person}{Luca Benini}, {and} \bibinfo{person}{Simone Benatti}.} \bibinfo{year}{2022}\natexlab{b}.
\newblock \showarticletitle{Embedding Temporal Convolutional Networks for Energy-efficient PPG-based Heart Rate Monitoring}.
\newblock \bibinfo{journal}{\emph{ACM Trans. Comput. Healthcare}} \bibinfo{volume}{3}, \bibinfo{number}{2}, Article \bibinfo{articleno}{19} (\bibinfo{date}{mar} \bibinfo{year}{2022}), \bibinfo{numpages}{25}~pages.
\newblock
\href{https://doi.org/10.1145/3487910}{doi:\nolinkurl{10.1145/3487910}}


\bibitem[Burrello et~al\mbox{.}(2021)]%
        {q-ppg}
\bibfield{author}{\bibinfo{person}{Alessio Burrello}, \bibinfo{person}{Daniele~Jahier Pagliari}, \bibinfo{person}{Matteo Risso}, \bibinfo{person}{Simone Benatti}, \bibinfo{person}{Enrico Macii}, \bibinfo{person}{Luca Benini}, {and} \bibinfo{person}{Massimo Poncino}.} \bibinfo{year}{2021}\natexlab{}.
\newblock \showarticletitle{Q-PPG: Energy-Efficient PPG-Based Heart Rate Monitoring on Wearable Devices}.
\newblock \bibinfo{journal}{\emph{IEEE Transactions on Biomedical Circuits and Systems}} \bibinfo{volume}{15}, \bibinfo{number}{6} (\bibinfo{year}{2021}), \bibinfo{pages}{1196--1209}.
\newblock
\href{https://doi.org/10.1109/TBCAS.2021.3122017}{doi:\nolinkurl{10.1109/TBCAS.2021.3122017}}


\bibitem[Cai and Vasconcelos(2020)]%
        {cai2020rethinkingdifferentiablesearchmixedprecision}
\bibfield{author}{\bibinfo{person}{Zhaowei Cai} {and} \bibinfo{person}{Nuno Vasconcelos}.} \bibinfo{year}{2020}\natexlab{}.
\newblock \bibinfo{title}{Rethinking Differentiable Search for Mixed-Precision Neural Networks}.
\newblock
\showeprint[arxiv]{2004.05795}~[cs.LG]
\urldef\tempurl%
\url{https://arxiv.org/abs/2004.05795}
\showURL{%
\tempurl}


\bibitem[Carlson et~al\mbox{.}(2021)]%
        {BCG_dataset}
\bibfield{author}{\bibinfo{person}{Charles Carlson}, \bibinfo{person}{Vanessa-Rose Turpin}, \bibinfo{person}{Ahmad Suliman}, \bibinfo{person}{Carl Ade}, \bibinfo{person}{Steve Warren}, {and} \bibinfo{person}{David~E. Thompson}.} \bibinfo{year}{2021}\natexlab{}.
\newblock \showarticletitle{Bed-Based Ballistocardiography: Dataset and Ability to Track Cardiovascular Parameters}.
\newblock \bibinfo{journal}{\emph{Sensors}} \bibinfo{volume}{21}, \bibinfo{number}{1} (\bibinfo{year}{2021}).
\newblock
\showISSN{1424-8220}
\href{https://doi.org/10.3390/s21010156}{doi:\nolinkurl{10.3390/s21010156}}


\bibitem[Cheng et~al\mbox{.}(2021)]%
        {ppg-dnn-2}
\bibfield{author}{\bibinfo{person}{Juan Cheng}, \bibinfo{person}{Yufei Xu}, \bibinfo{person}{Rencheng Song}, \bibinfo{person}{Yu Liu}, \bibinfo{person}{Chang Li}, {and} \bibinfo{person}{Xun Chen}.} \bibinfo{year}{2021}\natexlab{}.
\newblock \showarticletitle{Prediction of arterial blood pressure waveforms from photoplethysmogram signals via fully convolutional neural networks}.
\newblock \bibinfo{journal}{\emph{Computers in Biology and Medicine}}  \bibinfo{volume}{138} (\bibinfo{date}{11} \bibinfo{year}{2021}), \bibinfo{pages}{104877}.
\newblock
\showISSN{0010-4825}
\href{https://doi.org/10.1016/J.COMPBIOMED.2021.104877}{doi:\nolinkurl{10.1016/J.COMPBIOMED.2021.104877}}


\bibitem[Cherry et~al\mbox{.}(2025)]%
        {rt-ppg-id}
\bibfield{author}{\bibinfo{person}{Ali Cherry}, \bibinfo{person}{Aya Nasser}, \bibinfo{person}{Wassim Salameh}, \bibinfo{person}{Mohamad Abou~Ali}, {and} \bibinfo{person}{Mohamad Hajj-Hassan}.} \bibinfo{year}{2025}\natexlab{}.
\newblock \showarticletitle{Real-{Time} {PPG}-{Based} {Biometric} {Identification}: {Advancing} {Security} with {2D} {Gram} {Matrices} and {Deep} {Learning} {Models}}.
\newblock \bibinfo{journal}{\emph{Sensors}} \bibinfo{volume}{25}, \bibinfo{number}{1} (\bibinfo{date}{Jan.} \bibinfo{year}{2025}), \bibinfo{pages}{40}.
\newblock
\showISSN{1424-8220}
\href{https://doi.org/10.3390/s25010040}{doi:\nolinkurl{10.3390/s25010040}}
\newblock
\shownote{Publisher: Multidisciplinary Digital Publishing Institute}.


\bibitem[Choi et~al\mbox{.}(2018)]%
        {pact}
\bibfield{author}{\bibinfo{person}{Jungwook Choi}, \bibinfo{person}{Zhuo Wang}, \bibinfo{person}{Swagath Venkataramani}, \bibinfo{person}{Pierce~I{-}Jen Chuang}, \bibinfo{person}{Vijayalakshmi Srinivasan}, {and} \bibinfo{person}{Kailash Gopalakrishnan}.} \bibinfo{year}{2018}\natexlab{}.
\newblock \showarticletitle{{PACT:} Parameterized Clipping Activation for Quantized Neural Networks}.
\newblock \bibinfo{journal}{\emph{CoRR}}  \bibinfo{volume}{abs/1805.06085} (\bibinfo{year}{2018}).
\newblock
\showeprint[arXiv]{1805.06085}
\urldef\tempurl%
\url{http://arxiv.org/abs/1805.06085}
\showURL{%
\tempurl}


\bibitem[Costa et~al\mbox{.}(2023)]%
        {10147220}
\bibfield{author}{\bibinfo{person}{Thiago Bulhões Da~Silva Costa}, \bibinfo{person}{Felipe~Meneguitti Dias}, \bibinfo{person}{Diego Armando~Cardona Cardenas}, \bibinfo{person}{Marcelo Arruda Fiuza~De Toledo}, \bibinfo{person}{Daniel Mário~De Lima}, \bibinfo{person}{Jose~Eduardo Krieger}, {and} \bibinfo{person}{Marco~Antonio Gutierrez}.} \bibinfo{year}{2023}\natexlab{}.
\newblock \showarticletitle{Blood Pressure Estimation From Photoplethysmography by Considering Intra- and Inter-Subject Variabilities: Guidelines for a Fair Assessment}.
\newblock \bibinfo{journal}{\emph{IEEE Access}}  \bibinfo{volume}{11} (\bibinfo{year}{2023}), \bibinfo{pages}{57934--57950}.
\newblock
\href{https://doi.org/10.1109/ACCESS.2023.3284458}{doi:\nolinkurl{10.1109/ACCESS.2023.3284458}}


\bibitem[El~Hajj and Kyriacou(2020)]%
        {ppg-dnn-1}
\bibfield{author}{\bibinfo{person}{Chadi El~Hajj} {and} \bibinfo{person}{Panayiotis~A. Kyriacou}.} \bibinfo{year}{2020}\natexlab{}.
\newblock \showarticletitle{Cuffless and Continuous Blood Pressure Estimation From PPG Signals Using Recurrent Neural Networks}. In \bibinfo{booktitle}{\emph{2020 42nd Annual International Conference of the IEEE Engineering in Medicine \& Biology Society (EMBC)}}. \bibinfo{pages}{4269--4272}.
\newblock
\href{https://doi.org/10.1109/EMBC44109.2020.9175699}{doi:\nolinkurl{10.1109/EMBC44109.2020.9175699}}


\bibitem[Fong et~al\mbox{.}(2019)]%
        {ppg-svr}
\bibfield{author}{\bibinfo{person}{Mark Wong~Kei Fong}, \bibinfo{person}{E.~Y.K. Ng}, \bibinfo{person}{Kenneth Er~Zi Jian}, {and} \bibinfo{person}{Tan~Jen Hong}.} \bibinfo{year}{2019}\natexlab{}.
\newblock \showarticletitle{SVR ensemble-based continuous blood pressure prediction using multi-channel photoplethysmogram}.
\newblock \bibinfo{journal}{\emph{Computers in Biology and Medicine}}  \bibinfo{volume}{113} (\bibinfo{date}{10} \bibinfo{year}{2019}), \bibinfo{pages}{103392}.
\newblock
\showISSN{0010-4825}
\href{https://doi.org/10.1016/J.COMPBIOMED.2019.103392}{doi:\nolinkurl{10.1016/J.COMPBIOMED.2019.103392}}


\bibitem[Fuchs and Whelton(2020)]%
        {doi:10.1161/HYPERTENSIONAHA.119.14240}
\bibfield{author}{\bibinfo{person}{Flávio~D. Fuchs} {and} \bibinfo{person}{Paul~K. Whelton}.} \bibinfo{year}{2020}\natexlab{}.
\newblock \showarticletitle{High Blood Pressure and Cardiovascular Disease}.
\newblock \bibinfo{journal}{\emph{Hypertension}} \bibinfo{volume}{75}, \bibinfo{number}{2} (\bibinfo{year}{2020}), \bibinfo{pages}{285--292}.
\newblock
\showeprint{https://www.ahajournals.org/doi/pdf/10.1161/HYPERTENSIONAHA.119.14240}
\href{https://doi.org/10.1161/HYPERTENSIONAHA.119.14240}{doi:\nolinkurl{10.1161/HYPERTENSIONAHA.119.14240}}


\bibitem[Garofalo et~al\mbox{.}(2019)]%
        {pulp-nn}
\bibfield{author}{\bibinfo{person}{Angelo Garofalo}, \bibinfo{person}{Manuele Rusci}, \bibinfo{person}{Francesco Conti}, \bibinfo{person}{Davide Rossi}, {and} \bibinfo{person}{Luca Benini}.} \bibinfo{year}{2019}\natexlab{}.
\newblock \showarticletitle{{PULP-NN:} Accelerating Quantized Neural Networks on Parallel Ultra-Low-Power {RISC-V} Processors}.
\newblock \bibinfo{journal}{\emph{CoRR}}  \bibinfo{volume}{abs/1908.11263} (\bibinfo{year}{2019}).
\newblock
\showeprint[arXiv]{1908.11263}
\urldef\tempurl%
\url{http://arxiv.org/abs/1908.11263}
\showURL{%
\tempurl}


\bibitem[Geerse et~al\mbox{.}(2019)]%
        {Geerse2019}
\bibfield{author}{\bibinfo{person}{Carlijn Geerse}, \bibinfo{person}{Cher van Slobbe}, \bibinfo{person}{Edda van Triet}, {and} \bibinfo{person}{Lianne Simonse}.} \bibinfo{year}{2019}\natexlab{}.
\newblock \showarticletitle{Design of a care pathway for preventive blood pressure monitoring: Qualitative study}.
\newblock \bibinfo{journal}{\emph{JMIR Cardio}}  \bibinfo{volume}{3} (\bibinfo{date}{1} \bibinfo{year}{2019}).
\newblock
Issue 1.
\showISSN{25611011}
\href{https://doi.org/10.2196/13048}{doi:\nolinkurl{10.2196/13048}}


\bibitem[González et~al\mbox{.}(2023)]%
        {nature-survey}
\bibfield{author}{\bibinfo{person}{Sergio González}, \bibinfo{person}{Wan~Ting Hsieh}, {and} \bibinfo{person}{Trista Pei~Chun Chen}.} \bibinfo{year}{2023}\natexlab{}.
\newblock \showarticletitle{A benchmark for machine-learning based non-invasive blood pressure estimation using photoplethysmogram}.
\newblock \bibinfo{journal}{\emph{Scientific Data 2023 10:1}}  \bibinfo{volume}{10} (\bibinfo{date}{3} \bibinfo{year}{2023}), \bibinfo{pages}{1--16}.
\newblock
Issue 1.
\showISBNx{191.07/100.67}
\showISSN{2052-4463}
\href{https://doi.org/10.1038/s41597-023-02020-6}{doi:\nolinkurl{10.1038/s41597-023-02020-6}}


\bibitem[He et~al\mbox{.}(2016)]%
        {ppg-rf}
\bibfield{author}{\bibinfo{person}{Rui He}, \bibinfo{person}{Zhi-Pei Huang}, \bibinfo{person}{Lian-Ying Ji}, \bibinfo{person}{Jian-Kang Wu}, \bibinfo{person}{Huihui Li}, {and} \bibinfo{person}{Zhi-Qiang Zhang}.} \bibinfo{year}{2016}\natexlab{}.
\newblock \showarticletitle{Beat-to-beat ambulatory blood pressure estimation based on random forest}. In \bibinfo{booktitle}{\emph{2016 IEEE 13th International Conference on Wearable and Implantable Body Sensor Networks (BSN)}}. \bibinfo{pages}{194--198}.
\newblock
\href{https://doi.org/10.1109/BSN.2016.7516258}{doi:\nolinkurl{10.1109/BSN.2016.7516258}}


\bibitem[Howard et~al\mbox{.}(2017)]%
        {mobilenet}
\bibfield{author}{\bibinfo{person}{Andrew~G. Howard}, \bibinfo{person}{Menglong Zhu}, \bibinfo{person}{Bo Chen}, \bibinfo{person}{Dmitry Kalenichenko}, \bibinfo{person}{Weijun Wang}, \bibinfo{person}{Tobias Weyand}, \bibinfo{person}{Marco Andreetto}, {and} \bibinfo{person}{Hartwig Adam}.} \bibinfo{year}{2017}\natexlab{}.
\newblock \showarticletitle{MobileNets: Efficient Convolutional Neural Networks for Mobile Vision Applications}.
\newblock \bibinfo{journal}{\emph{CoRR}}  \bibinfo{volume}{abs/1704.04861} (\bibinfo{year}{2017}).
\newblock
\showeprint[arXiv]{1704.04861}
\urldef\tempurl%
\url{http://arxiv.org/abs/1704.04861}
\showURL{%
\tempurl}


\bibitem[Huang et~al\mbox{.}(2022)]%
        {ppg-mlp}
\bibfield{author}{\bibinfo{person}{Bin Huang}, \bibinfo{person}{Weihai Chen}, \bibinfo{person}{Chun~Liang Lin}, \bibinfo{person}{Chia~Feng Juang}, {and} \bibinfo{person}{Jianhua Wang}.} \bibinfo{year}{2022}\natexlab{}.
\newblock \showarticletitle{MLP-BP: A novel framework for cuffless blood pressure measurement with PPG and ECG signals based on MLP-Mixer neural networks}.
\newblock \bibinfo{journal}{\emph{Biomedical Signal Processing and Control}}  \bibinfo{volume}{73} (\bibinfo{date}{3} \bibinfo{year}{2022}), \bibinfo{pages}{103404}.
\newblock
\showISSN{1746-8094}
\href{https://doi.org/10.1016/J.BSPC.2021.103404}{doi:\nolinkurl{10.1016/J.BSPC.2021.103404}}


\bibitem[iMDsoft MetaVision~ICU(2025)]%
        {imdsoft-metavision-icu}
\bibfield{author}{\bibinfo{person}{iMDsoft MetaVision~ICU}.} \bibinfo{year}{Access 27/03/2025}\natexlab{}.
\newblock \bibinfo{title}{iMDsoft MetaVision ICU, https://imd-soft.com/metavision/icu/}.
\newblock


\bibitem[Jagannathan et~al\mbox{.}(2019)]%
        {PMID:31222515}
\bibfield{author}{\bibinfo{person}{Ram Jagannathan}, \bibinfo{person}{Shivani~A Patel}, \bibinfo{person}{Mohammed~K Ali}, {and} \bibinfo{person}{K~M~Venkat Narayan}.} \bibinfo{year}{2019}\natexlab{}.
\newblock \showarticletitle{Global Updates on Cardiovascular Disease Mortality Trends and Attribution of Traditional Risk Factors}.
\newblock \bibinfo{journal}{\emph{Current diabetes reports}} \bibinfo{volume}{19}, \bibinfo{number}{7} (\bibinfo{date}{June} \bibinfo{year}{2019}), \bibinfo{pages}{44}.
\newblock
\showISSN{1534-4827}
\href{https://doi.org/10.1007/s11892-019-1161-2}{doi:\nolinkurl{10.1007/s11892-019-1161-2}}


\bibitem[{Jahier Pagliari} et~al\mbox{.}(2023)]%
        {plinio}
\bibfield{author}{\bibinfo{person}{D. {Jahier Pagliari}}, \bibinfo{person}{M. {Risso}}, \bibinfo{person}{B.~A. {Motetti}}, {and} \bibinfo{person}{A. {Burrello}}.} \bibinfo{year}{2023}\natexlab{}.
\newblock \bibinfo{title}{PLiNIO: A User-Friendly Library of Gradient-based Methods for Complexity-aware DNN Optimization}.
\newblock
\showeprint[arxiv]{2307.09488}~[cs.LG]


\bibitem[Johnson et~al\mbox{.}(2016)]%
        {mimic_iii}
\bibfield{author}{\bibinfo{person}{Alistair~E.W. Johnson}, \bibinfo{person}{Tom~J. Pollard}, \bibinfo{person}{Lu Shen}, \bibinfo{person}{Li~Wei~H. Lehman}, \bibinfo{person}{Mengling Feng}, \bibinfo{person}{Mohammad Ghassemi}, \bibinfo{person}{Benjamin Moody}, \bibinfo{person}{Peter Szolovits}, \bibinfo{person}{Leo~Anthony Celi}, {and} \bibinfo{person}{Roger~G. Mark}.} \bibinfo{year}{2016}\natexlab{}.
\newblock \showarticletitle{MIMIC-III, a freely accessible critical care database}.
\newblock \bibinfo{journal}{\emph{Scientific Data}}  \bibinfo{volume}{3} (\bibinfo{date}{5} \bibinfo{year}{2016}), \bibinfo{pages}{1--9}.
\newblock
Issue 1.
\showISSN{20524463}
\href{https://doi.org/10.1038/SDATA.2016.35;SUBJMETA=139,1750,308,409,692,700;KWRD=DIAGNOSIS,HEALTH+CARE,MEDICAL+RESEARCH,OUTCOMES+RESEARCH,PROGNOSIS}{doi:\nolinkurl{10.1038/SDATA.2016.35;SUBJMETA=139,1750,308,409,692,700;KWRD=DIAGNOSIS,HEALTH+CARE,MEDICAL+RESEARCH,OUTCOMES+RESEARCH,PROGNOSIS}}


\bibitem[Joseph et~al\mbox{.}(2014)]%
        {ppg-signal}
\bibfield{author}{\bibinfo{person}{Greeshma Joseph}, \bibinfo{person}{Almaria Joseph}, \bibinfo{person}{Geevarghese Titus}, \bibinfo{person}{Rintu~Mariya Thomas}, {and} \bibinfo{person}{Dency Jose}.} \bibinfo{year}{2014}\natexlab{}.
\newblock \showarticletitle{Photoplethysmogram (PPG) signal analysis and wavelet de-noising}. In \bibinfo{booktitle}{\emph{2014 Annual International Conference on Emerging Research Areas: Magnetics, Machines and Drives (AICERA/iCMMD)}}. \bibinfo{pages}{1--5}.
\newblock
\href{https://doi.org/10.1109/AICERA.2014.6908199}{doi:\nolinkurl{10.1109/AICERA.2014.6908199}}


\bibitem[Joseph and T.S(2024)]%
        {10.1145/3699512}
\bibfield{author}{\bibinfo{person}{Tresa Joseph} {and} \bibinfo{person}{Bindiya T.S}.} \bibinfo{year}{2024}\natexlab{}.
\newblock \showarticletitle{Real-time Blood Pressure Prediction on Wearables with Edge-Based DNNs: A Co-Design Approach}.
\newblock \bibinfo{journal}{\emph{ACM Trans. Des. Autom. Electron. Syst.}} \bibinfo{volume}{30}, \bibinfo{number}{1}, Article \bibinfo{articleno}{11} (\bibinfo{date}{Dec.} \bibinfo{year}{2024}), \bibinfo{numpages}{24}~pages.
\newblock
\showISSN{1084-4309}
\href{https://doi.org/10.1145/3699512}{doi:\nolinkurl{10.1145/3699512}}


\bibitem[Kachuee et~al\mbox{.}(2015)]%
        {uci_dataset}
\bibfield{author}{\bibinfo{person}{Mohamad Kachuee}, \bibinfo{person}{Mohammad~Mahdi Kiani}, \bibinfo{person}{Hoda Mohammadzade}, {and} \bibinfo{person}{Mahdi Shabany}.} \bibinfo{year}{2015}\natexlab{}.
\newblock \showarticletitle{Cuff-less high-accuracy calibration-free blood pressure estimation using pulse transit time}. In \bibinfo{booktitle}{\emph{2015 IEEE International Symposium on Circuits and Systems (ISCAS)}}. \bibinfo{pages}{1006--1009}.
\newblock
\href{https://doi.org/10.1109/ISCAS.2015.7168806}{doi:\nolinkurl{10.1109/ISCAS.2015.7168806}}


\bibitem[Kachuee et~al\mbox{.}(2017)]%
        {cuffless2017}
\bibfield{author}{\bibinfo{person}{Mohammad Kachuee}, \bibinfo{person}{Mohammad~Mahdi Kiani}, \bibinfo{person}{Hoda Mohammadzade}, {and} \bibinfo{person}{Mahdi Shabany}.} \bibinfo{year}{2017}\natexlab{}.
\newblock \showarticletitle{Cuffless Blood Pressure Estimation Algorithms for Continuous Health-Care Monitoring}.
\newblock \bibinfo{journal}{\emph{IEEE Transactions on Biomedical Engineering}} \bibinfo{volume}{64}, \bibinfo{number}{4} (\bibinfo{year}{2017}), \bibinfo{pages}{859--869}.
\newblock
\href{https://doi.org/10.1109/TBME.2016.2580904}{doi:\nolinkurl{10.1109/TBME.2016.2580904}}


\bibitem[Le et~al\mbox{.}(2020)]%
        {9268175}
\bibfield{author}{\bibinfo{person}{Tai Le}, \bibinfo{person}{Floranne Ellington}, \bibinfo{person}{Tao-Yi Lee}, \bibinfo{person}{Khuong Vo}, \bibinfo{person}{Michelle Khine}, \bibinfo{person}{Sandeep~Kumar Krishnan}, \bibinfo{person}{Nikil Dutt}, {and} \bibinfo{person}{Hung Cao}.} \bibinfo{year}{2020}\natexlab{}.
\newblock \showarticletitle{Continuous Non-Invasive Blood Pressure Monitoring: A Methodological Review on Measurement Techniques}.
\newblock \bibinfo{journal}{\emph{IEEE Access}}  \bibinfo{volume}{8} (\bibinfo{year}{2020}), \bibinfo{pages}{212478--212498}.
\newblock
\href{https://doi.org/10.1109/ACCESS.2020.3040257}{doi:\nolinkurl{10.1109/ACCESS.2020.3040257}}


\bibitem[Lee et~al\mbox{.}(2024)]%
        {10568937}
\bibfield{author}{\bibinfo{person}{Nayoung Lee}, \bibinfo{person}{Sang-Hyun Kim}, \bibinfo{person}{Misoon Lee}, {and} \bibinfo{person}{Jiyoung Woo}.} \bibinfo{year}{2024}\natexlab{}.
\newblock \showarticletitle{Advancing Continuous Blood Pressure Estimation with Transformer on Photoplethysmography in Operation Room}.
\newblock \bibinfo{journal}{\emph{IEEE Access}}  \bibinfo{volume}{12} (\bibinfo{year}{2024}), \bibinfo{pages}{90486--90500}.
\newblock
\href{https://doi.org/10.1109/ACCESS.2024.3417940}{doi:\nolinkurl{10.1109/ACCESS.2024.3417940}}


\bibitem[Li et~al\mbox{.}(2023)]%
        {ppg_auth}
\bibfield{author}{\bibinfo{person}{Lin Li}, \bibinfo{person}{Chao Chen}, \bibinfo{person}{Lei Pan}, \bibinfo{person}{Leo~Yu Zhang}, \bibinfo{person}{Zhifeng Wang}, \bibinfo{person}{Jun Zhang}, {and} \bibinfo{person}{Yang Xiang}.} \bibinfo{year}{2023}\natexlab{}.
\newblock \showarticletitle{A {Survey} of {PPG}'s {Application} in {Authentication}}.
\newblock \bibinfo{journal}{\emph{Computers \& Security}}  \bibinfo{volume}{135} (\bibinfo{date}{Dec.} \bibinfo{year}{2023}), \bibinfo{pages}{103488}.
\newblock
\showISSN{01674048}
\href{https://doi.org/10.1016/j.cose.2023.103488}{doi:\nolinkurl{10.1016/j.cose.2023.103488}}


\bibitem[Li et~al\mbox{.}(2020)]%
        {s20195606}
\bibfield{author}{\bibinfo{person}{Yung-Hui Li}, \bibinfo{person}{Latifa~Nabila Harfiya}, \bibinfo{person}{Kartika Purwandari}, {and} \bibinfo{person}{Yue-Der Lin}.} \bibinfo{year}{2020}\natexlab{}.
\newblock \showarticletitle{Real-Time Cuffless Continuous Blood Pressure Estimation Using Deep Learning Model}.
\newblock \bibinfo{journal}{\emph{Sensors}} \bibinfo{volume}{20}, \bibinfo{number}{19} (\bibinfo{year}{2020}).
\newblock
\showISSN{1424-8220}
\href{https://doi.org/10.3390/s20195606}{doi:\nolinkurl{10.3390/s20195606}}


\bibitem[Liang et~al\mbox{.}(2018)]%
        {PPGB_dataset}
\bibfield{author}{\bibinfo{person}{Yongbo Liang}, \bibinfo{person}{Zhencheng Chen}, \bibinfo{person}{Guiyong Liu}, {and} \bibinfo{person}{Mohamed Elgendi}.} \bibinfo{year}{2018}\natexlab{}.
\newblock \showarticletitle{A new, short-recorded photoplethysmogram dataset for blood pressure monitoring in China}.
\newblock \bibinfo{journal}{\emph{Scientific Data}}  \bibinfo{volume}{5} (\bibinfo{date}{2} \bibinfo{year}{2018}).
\newblock
\href{https://doi.org/10.1038/sdata.2018.20}{doi:\nolinkurl{10.1038/sdata.2018.20}}


\bibitem[Lim et~al\mbox{.}(2025)]%
        {Lim2025}
\bibfield{author}{\bibinfo{person}{Sungjun Lim}, \bibinfo{person}{Taero Kim}, \bibinfo{person}{Hyeonjeong Lee}, \bibinfo{person}{Yewon Kim}, \bibinfo{person}{Minhoi Park}, \bibinfo{person}{Kwang~Yong Kim}, \bibinfo{person}{Minseong Kim}, \bibinfo{person}{Kyu~Hyung Kim}, \bibinfo{person}{Jiyoung Jung}, {and} \bibinfo{person}{Kyungwoo Song}.} \bibinfo{year}{2025}\natexlab{}.
\newblock \showarticletitle{Robust optimization for PPG-based blood pressure estimation}.
\newblock \bibinfo{journal}{\emph{Biomedical Signal Processing and Control}}  \bibinfo{volume}{105} (\bibinfo{date}{7} \bibinfo{year}{2025}), \bibinfo{pages}{107585}.
\newblock
\showISSN{1746-8094}
\href{https://doi.org/10.1016/J.BSPC.2025.107585}{doi:\nolinkurl{10.1016/J.BSPC.2025.107585}}


\bibitem[Lin et~al\mbox{.}(2021)]%
        {9630488}
\bibfield{author}{\bibinfo{person}{Wenrui Lin}, \bibinfo{person}{Berken~Utku Demirel}, \bibinfo{person}{Mohammad~Abdullah Al~Faruque}, {and} \bibinfo{person}{G.P. Li}.} \bibinfo{year}{2021}\natexlab{}.
\newblock \showarticletitle{Energy-efficient Blood Pressure Monitoring based on Single-site Photoplethysmogram on Wearable Devices}. In \bibinfo{booktitle}{\emph{2021 43rd Annual International Conference of the IEEE Engineering in Medicine \& Biology Society (EMBC)}}. \bibinfo{pages}{504--507}.
\newblock
\href{https://doi.org/10.1109/EMBC46164.2021.9630488}{doi:\nolinkurl{10.1109/EMBC46164.2021.9630488}}


\bibitem[Liu et~al\mbox{.}(2018)]%
        {darts}
\bibfield{author}{\bibinfo{person}{Hanxiao Liu}, \bibinfo{person}{Karen Simonyan}, {and} \bibinfo{person}{Yiming Yang}.} \bibinfo{year}{2018}\natexlab{}.
\newblock \showarticletitle{{DARTS:} Differentiable Architecture Search}.
\newblock \bibinfo{journal}{\emph{CoRR}}  \bibinfo{volume}{abs/1806.09055} (\bibinfo{year}{2018}).
\newblock
\showeprint[arXiv]{1806.09055}
\urldef\tempurl%
\url{http://arxiv.org/abs/1806.09055}
\showURL{%
\tempurl}


\bibitem[Luepker et~al\mbox{.}(6 01)]%
        {ctx2426934390007866}
\bibfield{author}{\bibinfo{person}{Russell~V Luepker}, \bibinfo{person}{Donna~K Arnett}, \bibinfo{person}{David~R Jacobs}, \bibinfo{person}{Susan~J Duval}, \bibinfo{person}{Aaron~R Folsom}, \bibinfo{person}{Christopher Armstrong}, {and} \bibinfo{person}{Henry Blackburn}.} \bibinfo{year}{2006-01}\natexlab{}.
\newblock \showarticletitle{Trends in Blood Pressure, Hypertension Control, and Stroke Mortality: The Minnesota Heart Survey}.
\newblock \bibinfo{journal}{\emph{The American journal of medicine.}} \bibinfo{volume}{119}, \bibinfo{number}{1} (\bibinfo{year}{2006-01}).
\newblock
\showISSN{0002-9343}
\showLCCN{2004700037}


\bibitem[Martínez et~al\mbox{.}(2018)]%
        {correlation_study}
\bibfield{author}{\bibinfo{person}{Gloria Martínez}, \bibinfo{person}{Newton Howard}, \bibinfo{person}{Derek Abbott}, \bibinfo{person}{Kenneth Lim}, \bibinfo{person}{Rabab Ward}, {and} \bibinfo{person}{Mohamed Elgendi}.} \bibinfo{year}{2018}\natexlab{}.
\newblock \showarticletitle{Can Photoplethysmography Replace Arterial Blood Pressure in the Assessment of Blood Pressure?}
\newblock \bibinfo{journal}{\emph{Journal of Clinical Medicine}} \bibinfo{volume}{7}, \bibinfo{number}{10} (\bibinfo{year}{2018}).
\newblock
\showISSN{2077-0383}
\href{https://doi.org/10.3390/jcm7100316}{doi:\nolinkurl{10.3390/jcm7100316}}


\bibitem[Meidert and Saugel(2017)]%
        {Meidert2017}
\bibfield{author}{\bibinfo{person}{Agnes~S. Meidert} {and} \bibinfo{person}{Bernd Saugel}.} \bibinfo{year}{2017}\natexlab{}.
\newblock \bibinfo{title}{Techniques for non-invasive monitoring of arterial blood pressure}.
\newblock
Issue JAN.
\showISSN{2296858X}
\href{https://doi.org/10.3389/fmed.2017.00231}{doi:\nolinkurl{10.3389/fmed.2017.00231}}


\bibitem[Motetti et~al\mbox{.}(2024)]%
        {10644100}
\bibfield{author}{\bibinfo{person}{Beatrice~Alessandra Motetti}, \bibinfo{person}{Matteo Risso}, \bibinfo{person}{Alessio Burrello}, \bibinfo{person}{Enrico Macii}, \bibinfo{person}{Massimo Poncino}, {and} \bibinfo{person}{Daniele~Jahier Pagliari}.} \bibinfo{year}{2024}\natexlab{}.
\newblock \showarticletitle{Joint Pruning and Channel-Wise Mixed-Precision Quantization for Efficient Deep Neural Networks}.
\newblock \bibinfo{journal}{\emph{IEEE Trans. Comput.}} \bibinfo{volume}{73}, \bibinfo{number}{11} (\bibinfo{year}{2024}), \bibinfo{pages}{2619--2633}.
\newblock
\href{https://doi.org/10.1109/TC.2024.3449084}{doi:\nolinkurl{10.1109/TC.2024.3449084}}


\bibitem[Mukkamala et~al\mbox{.}(2015)]%
        {7118672}
\bibfield{author}{\bibinfo{person}{Ramakrishna Mukkamala}, \bibinfo{person}{Jin-Oh Hahn}, \bibinfo{person}{Omer~T. Inan}, \bibinfo{person}{Lalit~K. Mestha}, \bibinfo{person}{Chang-Sei Kim}, \bibinfo{person}{Hakan Töreyin}, {and} \bibinfo{person}{Survi Kyal}.} \bibinfo{year}{2015}\natexlab{}.
\newblock \showarticletitle{Toward Ubiquitous Blood Pressure Monitoring via Pulse Transit Time: Theory and Practice}.
\newblock \bibinfo{journal}{\emph{IEEE Transactions on Biomedical Engineering}} \bibinfo{volume}{62}, \bibinfo{number}{8} (\bibinfo{year}{2015}), \bibinfo{pages}{1879--1901}.
\newblock
\href{https://doi.org/10.1109/TBME.2015.2441951}{doi:\nolinkurl{10.1109/TBME.2015.2441951}}


\bibitem[Odutayo et~al\mbox{.}(2016)]%
        {Odutayoi4482}
\bibfield{author}{\bibinfo{person}{Ayodele Odutayo}, \bibinfo{person}{Christopher~X Wong}, \bibinfo{person}{Allan~J Hsiao}, \bibinfo{person}{Sally Hopewell}, \bibinfo{person}{Douglas~G Altman}, {and} \bibinfo{person}{Connor~A Emdin}.} \bibinfo{year}{2016}\natexlab{}.
\newblock \showarticletitle{Atrial fibrillation and risks of cardiovascular disease, renal disease, and death: systematic review and meta-analysis}.
\newblock \bibinfo{journal}{\emph{BMJ}}  \bibinfo{volume}{354} (\bibinfo{year}{2016}).
\newblock
\showeprint{https://www.bmj.com/content/354/bmj.i4482.full.pdf}
\href{https://doi.org/10.1136/bmj.i4482}{doi:\nolinkurl{10.1136/bmj.i4482}}


\bibitem[Qin et~al\mbox{.}(2023)]%
        {resnet-ppg}
\bibfield{author}{\bibinfo{person}{Caijie Qin}, \bibinfo{person}{Yong Li}, \bibinfo{person}{Chibiao Liu}, {and} \bibinfo{person}{Xibo Ma}.} \bibinfo{year}{2023}\natexlab{}.
\newblock \showarticletitle{Cuff-Less Blood Pressure Prediction Based on Photoplethysmography and Modified ResNet}.
\newblock \bibinfo{journal}{\emph{Bioengineering}} \bibinfo{volume}{10}, \bibinfo{number}{4} (\bibinfo{year}{2023}).
\newblock
\showISSN{2306-5354}
\href{https://doi.org/10.3390/bioengineering10040400}{doi:\nolinkurl{10.3390/bioengineering10040400}}


\bibitem[Rishi~Vardhan et~al\mbox{.}(2021)]%
        {9680071}
\bibfield{author}{\bibinfo{person}{K Rishi~Vardhan}, \bibinfo{person}{S Vedanth}, \bibinfo{person}{G Poojah}, \bibinfo{person}{K Abhishek}, \bibinfo{person}{M Nitish~Kumar}, {and} \bibinfo{person}{Vineeth Vijayaraghavan}.} \bibinfo{year}{2021}\natexlab{}.
\newblock \showarticletitle{BP-Net: Efficient Deep Learning for Continuous Arterial Blood Pressure Estimation using Photoplethysmogram}. In \bibinfo{booktitle}{\emph{2021 20th IEEE International Conference on Machine Learning and Applications (ICMLA)}}. \bibinfo{pages}{1495--1500}.
\newblock
\href{https://doi.org/10.1109/ICMLA52953.2021.00241}{doi:\nolinkurl{10.1109/ICMLA52953.2021.00241}}


\bibitem[Ronneberger et~al\mbox{.}(2015)]%
        {unet}
\bibfield{author}{\bibinfo{person}{Olaf Ronneberger}, \bibinfo{person}{Philipp Fischer}, {and} \bibinfo{person}{Thomas Brox}.} \bibinfo{year}{2015}\natexlab{}.
\newblock \bibinfo{title}{U-Net: Convolutional Networks for Biomedical Image Segmentation}.
\newblock
\showeprint[arxiv]{1505.04597}~[cs.CV]


\bibitem[Roy et~al\mbox{.}(2022)]%
        {9913482}
\bibfield{author}{\bibinfo{person}{Monalisa~Singha Roy}, \bibinfo{person}{Rajarshi Gupta}, {and} \bibinfo{person}{Kaushik~Das Sharma}.} \bibinfo{year}{2022}\natexlab{}.
\newblock \showarticletitle{BePCon: A Photoplethysmography-Based Quality-Aware Continuous Beat-to-Beat Blood Pressure Measurement Technique Using Deep Learning}.
\newblock \bibinfo{journal}{\emph{IEEE Transactions on Instrumentation and Measurement}}  \bibinfo{volume}{71} (\bibinfo{year}{2022}), \bibinfo{pages}{1--9}.
\newblock
\href{https://doi.org/10.1109/TIM.2022.3212750}{doi:\nolinkurl{10.1109/TIM.2022.3212750}}


\bibitem[Saeed et~al\mbox{.}(2002)]%
        {mimic_ii}
\bibfield{author}{\bibinfo{person}{M. Saeed}, \bibinfo{person}{C. Lieu}, \bibinfo{person}{G. Raber}, {and} \bibinfo{person}{R.G. Mark}.} \bibinfo{year}{2002}\natexlab{}.
\newblock \showarticletitle{MIMIC II: a massive temporal ICU patient database to support research in intelligent patient monitoring}. In \bibinfo{booktitle}{\emph{Computers in Cardiology}}. \bibinfo{pages}{641--644}.
\newblock
\href{https://doi.org/10.1109/CIC.2002.1166854}{doi:\nolinkurl{10.1109/CIC.2002.1166854}}


\bibitem[Semiconductor(2024)]%
        {nordic-2}
\bibfield{author}{\bibinfo{person}{Nordic Semiconductor}.} \bibinfo{year}{Access 16/05/2024}\natexlab{}.
\newblock \bibinfo{title}{Nordic II, https://www.nordicsemi.com/}.
\newblock


\bibitem[Sheng et~al\mbox{.}(2024)]%
        {10798356}
\bibfield{author}{\bibinfo{person}{Mingda Sheng}, \bibinfo{person}{Rui Xing}, \bibinfo{person}{Youze Xin}, \bibinfo{person}{Bing Zhang}, \bibinfo{person}{Zhuoqi Guo}, \bibinfo{person}{Zhongming Xue}, {and} \bibinfo{person}{Li Geng}.} \bibinfo{year}{2024}\natexlab{}.
\newblock \showarticletitle{A 4.4$\mu$W Cuffless Blood Pressure Measurement Processor Based on Event-Driven and Module-Level Asynchronous Scheme}. In \bibinfo{booktitle}{\emph{2024 IEEE Biomedical Circuits and Systems Conference (BioCAS)}}. \bibinfo{pages}{1--5}.
\newblock
\href{https://doi.org/10.1109/BioCAS61083.2024.10798356}{doi:\nolinkurl{10.1109/BioCAS61083.2024.10798356}}


\bibitem[Stoller et~al\mbox{.}(2018)]%
        {stoller2018waveunetmultiscaleneuralnetwork}
\bibfield{author}{\bibinfo{person}{Daniel Stoller}, \bibinfo{person}{Sebastian Ewert}, {and} \bibinfo{person}{Simon Dixon}.} \bibinfo{year}{2018}\natexlab{}.
\newblock \bibinfo{title}{Wave-U-Net: A Multi-Scale Neural Network for End-to-End Audio Source Separation}.
\newblock
\showeprint[arxiv]{1806.03185}~[cs.SD]
\urldef\tempurl%
\url{https://arxiv.org/abs/1806.03185}
\showURL{%
\tempurl}


\bibitem[Sun et~al\mbox{.}(2023)]%
        {10.1145/3555776.3577747}
\bibfield{author}{\bibinfo{person}{Bailian Sun}, \bibinfo{person}{Safin Bayes}, \bibinfo{person}{Abdelrhman~Mohamed Abotaleb}, {and} \bibinfo{person}{Mohamed Hassan}.} \bibinfo{year}{2023}\natexlab{}.
\newblock \showarticletitle{The Case for tinyML in Healthcare: CNNs for Real-Time On-Edge Blood Pressure Estimation}. In \bibinfo{booktitle}{\emph{Proceedings of the 38th ACM/SIGAPP Symposium on Applied Computing}} (Tallinn, Estonia) \emph{(\bibinfo{series}{SAC '23})}. \bibinfo{publisher}{Association for Computing Machinery}, \bibinfo{address}{New York, NY, USA}, \bibinfo{pages}{629–638}.
\newblock
\showISBNx{9781450395175}
\href{https://doi.org/10.1145/3555776.3577747}{doi:\nolinkurl{10.1145/3555776.3577747}}


\bibitem[Technologies(2024)]%
        {gap8}
\bibfield{author}{\bibinfo{person}{Greenwaves Technologies}.} \bibinfo{year}{Access 16/05/2024}\natexlab{}.
\newblock \bibinfo{title}{Gap8, https://greenwaves-technologies.com/ gap8\_mcu\_ai/}.
\newblock


\bibitem[Tian et~al\mbox{.}(2025)]%
        {Tian2025}
\bibfield{author}{\bibinfo{person}{Zhonghe Tian}, \bibinfo{person}{Aiping Liu}, \bibinfo{person}{Guokang Zhu}, {and} \bibinfo{person}{Xun Chen}.} \bibinfo{year}{2025}\natexlab{}.
\newblock \showarticletitle{A paralleled CNN and Transformer network for PPG-based cuff-less blood pressure estimation}.
\newblock \bibinfo{journal}{\emph{Biomedical Signal Processing and Control}}  \bibinfo{volume}{99} (\bibinfo{date}{1} \bibinfo{year}{2025}), \bibinfo{pages}{106741}.
\newblock
\showISSN{1746-8094}
\href{https://doi.org/10.1016/J.BSPC.2024.106741}{doi:\nolinkurl{10.1016/J.BSPC.2024.106741}}


\bibitem[Vaswani et~al\mbox{.}(2017)]%
        {attention-is-all-you-need}
\bibfield{author}{\bibinfo{person}{Ashish Vaswani}, \bibinfo{person}{Noam Shazeer}, \bibinfo{person}{Niki Parmar}, \bibinfo{person}{Jakob Uszkoreit}, \bibinfo{person}{Llion Jones}, \bibinfo{person}{Aidan~N. Gomez}, \bibinfo{person}{\L{}ukasz Kaiser}, {and} \bibinfo{person}{Illia Polosukhin}.} \bibinfo{year}{2017}\natexlab{}.
\newblock \showarticletitle{Attention is all you need}. In \bibinfo{booktitle}{\emph{Proceedings of the 31st International Conference on Neural Information Processing Systems}} (Long Beach, California, USA) \emph{(\bibinfo{series}{NIPS'17})}. \bibinfo{publisher}{Curran Associates Inc.}, \bibinfo{address}{Red Hook, NY, USA}, \bibinfo{pages}{6000–6010}.
\newblock
\showISBNx{9781510860964}


\bibitem[Ward and Langton(2007)]%
        {Ward2007}
\bibfield{author}{\bibinfo{person}{Matthew Ward} {and} \bibinfo{person}{Jeremy~A. Langton}.} \bibinfo{year}{2007}\natexlab{}.
\newblock \showarticletitle{Blood pressure measurement}.
\newblock \bibinfo{journal}{\emph{Continuing Education in Anaesthesia, Critical Care and Pain}}  \bibinfo{volume}{7} (\bibinfo{year}{2007}), \bibinfo{pages}{122--126}.
\newblock
Issue 4.
\showISSN{17431824}
\href{https://doi.org/10.1093/bjaceaccp/mkm022}{doi:\nolinkurl{10.1093/bjaceaccp/mkm022}}


\bibitem[White et~al\mbox{.}(1993)]%
        {White1993}
\bibfield{author}{\bibinfo{person}{William~B. White}, \bibinfo{person}{Alan~S. Berson}, \bibinfo{person}{Carroll Robbins}, \bibinfo{person}{Michael~J. Jamieson}, \bibinfo{person}{L.~Michael Prisant}, \bibinfo{person}{Edward Roccella}, {and} \bibinfo{person}{Sheldon~G. Sheps}.} \bibinfo{year}{1993}\natexlab{}.
\newblock \showarticletitle{National standard for measurement of resting and ambulatory blood pressures with automated sphygmomanometers}.
\newblock \bibinfo{journal}{\emph{Hypertension (Dallas, Tex. : 1979)}}  \bibinfo{volume}{21} (\bibinfo{year}{1993}), \bibinfo{pages}{504--509}.
\newblock
Issue 4.
\showISSN{0194-911X}
\href{https://doi.org/10.1161/01.HYP.21.4.504}{doi:\nolinkurl{10.1161/01.HYP.21.4.504}}


\bibitem[Zhang et~al\mbox{.}(2022)]%
        {9871215}
\bibfield{author}{\bibinfo{person}{Jialong Zhang}, \bibinfo{person}{Jianzheng Li}, \bibinfo{person}{Yizhou Jiang}, \bibinfo{person}{Kai Wang}, \bibinfo{person}{Ran Guo}, \bibinfo{person}{Yu Ma}, {and} \bibinfo{person}{Yajie Qin}.} \bibinfo{year}{2022}\natexlab{}.
\newblock \showarticletitle{A Hardware-based Lightweight ANN for Real-time Wearable Blood Pressure Estimation}. In \bibinfo{booktitle}{\emph{2022 44th Annual International Conference of the IEEE Engineering in Medicine \& Biology Society (EMBC)}}. \bibinfo{pages}{4295--4298}.
\newblock
\href{https://doi.org/10.1109/EMBC48229.2022.9871215}{doi:\nolinkurl{10.1109/EMBC48229.2022.9871215}}


\bibitem[Zhang et~al\mbox{.}(2026)]%
        {Zhang2026}
\bibfield{author}{\bibinfo{person}{Yiming Zhang}, \bibinfo{person}{Shirong Qiu}, \bibinfo{person}{Kai Du}, \bibinfo{person}{Shun Wu}, \bibinfo{person}{Ting Xiang}, \bibinfo{person}{Kenghao Zheng}, \bibinfo{person}{Zijun Liu}, \bibinfo{person}{Hanjie Chen}, \bibinfo{person}{Nan Ji}, \bibinfo{person}{Fa Wang}, \bibinfo{person}{Weijia Wu}, {and} \bibinfo{person}{Yuan~Ting Zhang}.} \bibinfo{year}{2026}\natexlab{}.
\newblock \bibinfo{title}{Artificial Intelligence-Enhanced Wearable Blood Pressure Monitoring in Resource-Limited Settings: A Co-Design of Sensors, Model, and Deployment}.
\newblock
Issue 1.
\showISSN{21505551}
\href{https://doi.org/10.1007/s40820-025-02003-9}{doi:\nolinkurl{10.1007/s40820-025-02003-9}}


\bibitem[Zhang et~al\mbox{.}(2025)]%
        {10856555}
\bibfield{author}{\bibinfo{person}{Yiming Zhang}, \bibinfo{person}{Congcong Zhou}, \bibinfo{person}{Xianglin Ren}, \bibinfo{person}{Qing Wang}, \bibinfo{person}{Hongwei Wang}, \bibinfo{person}{Ting Xiang}, \bibinfo{person}{Shirong Qiu}, \bibinfo{person}{Yuan-Ting Zhang}, {and} \bibinfo{person}{Xuesong Ye}.} \bibinfo{year}{2025}\natexlab{}.
\newblock \showarticletitle{Personalized Continuous Blood Pressure Tracking Through Single Channel PPG in Wearable Scenarios}.
\newblock \bibinfo{journal}{\emph{IEEE Journal of Biomedical and Health Informatics}} \bibinfo{volume}{29}, \bibinfo{number}{6} (\bibinfo{year}{2025}), \bibinfo{pages}{4109--4120}.
\newblock
\href{https://doi.org/10.1109/JBHI.2025.3535788}{doi:\nolinkurl{10.1109/JBHI.2025.3535788}}


\bibitem[Zhou et~al\mbox{.}(2022)]%
        {ppg-tcn}
\bibfield{author}{\bibinfo{person}{Kai Zhou}, \bibinfo{person}{Zhixiang Yin}, \bibinfo{person}{Yu Peng}, {and} \bibinfo{person}{Zhiliang Zeng}.} \bibinfo{year}{2022}\natexlab{}.
\newblock \showarticletitle{Methods for Continuous Blood Pressure Estimation Using Temporal Convolutional Neural Networks and Ensemble Empirical Mode Decomposition}.
\newblock \bibinfo{journal}{\emph{Electronics}} \bibinfo{volume}{11}, \bibinfo{number}{9} (\bibinfo{year}{2022}).
\newblock
\showISSN{2079-9292}
\href{https://doi.org/10.3390/electronics11091378}{doi:\nolinkurl{10.3390/electronics11091378}}


\end{thebibliography}
\end{document}